\documentclass[pmlr]{jmlr}

\usepackage{makecell} 
\usepackage{multirow}
\usepackage[
]{sidecap}   
\sidecaptionvpos{figure}{t} 
\usepackage{longtable}
\usepackage{booktabs}
\usepackage[load-configurations=version-1]{siunitx} 

\theorembodyfont{\upshape}
\theoremheaderfont{\scshape}
\theorempostheader{:}
\theoremsep{\newline}
\usepackage{makecell} 
\usepackage{multirow}
\usepackage[ruled]{algorithm2e}
\usepackage{bm}
\usepackage[nooneline]{caption}

\title{Scale-Equivariant UNet for Histopathology Image Segmentation}

  \author{\Name{Yilong Yang} \Email{Yilong.Yang@soton.ac.uk}\\
  \Name{Srinandan Dasmahapatra} \Email{sd@ecs.soton.ac.uk}\\
  \Name{Sasan Mahmoodi} \Email{sm3@ecs.soton.ac.uk}\\
  \addr University of Southampton, University Road, Southampton, SO17 1BJ, United Kingdom}

\begin{document}

\maketitle

\begin{abstract}
Digital histopathology slides are scanned and viewed under different magnifications and stored as images at different resolutions. Convolutional Neural Networks (CNNs) trained on such images at a given scale fail to generalise to those at different scales.  This inability is often addressed by augmenting training data with re-scaled images, allowing a model with sufficient capacity to learn the requisite patterns. Alternatively, designing CNN filters to be scale-equivariant frees up model capacity to learn discriminative features. In this paper, we propose the Scale-Equivariant UNet (SEUNet) for image segmentation by building on scale-space theory. The SEUNet contains groups of filters that are linear combinations of Gaussian basis filters, whose scale parameters are trainable but constrained to span disjoint scales through the layers of the network. Extensive experiments on a nuclei segmentation dataset and a tissue type segmentation dataset demonstrate that our method outperforms other approaches, with much fewer trainable parameters.
\end{abstract}

\section{Introduction}
\label{sec:intro}
Pathologists diagnosing biopsy samples view histopathology slices at different magnifications by controlling the microscope's objective revolver. Neural network based decision support for digital pathology  take as input digital images scanned from glass slides. Specimen slides scanned at different medical institutions may use different objective magnifications to digitalize specimen slides, resulting in whole slide images (WSI) being at different scales. For example, images provided by the CRAG dataset \cite{awan2017glandular} are in 20$\times$ magnification; For the DigestPath-2019 dataset \cite{li2019signet}, images are in 40$\times$ magnification. Models such as Convolutional neural networks (CNNs) trained on images at a specific scale generally can not generalise to other scales, which greatly restricts the applicability of computer-aided diagnosis models.

CNNs have dominated the computer vision field since the proposal of the AlexNet \citep{krizhevsky2012imagenet}. The most widely adopted strategy to cope with scale variation in unseen data is introducing scale augmentation during training CNNs, where training samples are randomly scaled before being fed into the network. Other attempts such as scale selection \citep{girshick2014rich} and scale fusion \citep{kokkinos2015pushing} also help to circumvent scale changes. However, these methods lack explicit mechanisms to model scale information. Some works such as \cite{kanazawa2014locally,marcos2018scale,xu2014scale} achieve scale equivariance by resizing the input or filter, but these methods are computationally expensive since they rely on tensor resizing and image interpolation. Other ways of generating filters of different sizes include \cite{bekkers2019b,sosnovik2019scale, zhu2022scaling}, parameterising filters by a trainable linear combination of a family of predefined, fixed multi-scale basis functions (Hermite, Fourier, B-Splines). Such methods, however, require that both the scale of basis functions and the size of filters should be fixed, once the network has been initialised. The work presented in \citet{pintea2021resolution} shows that hard-coding the scale hyper-parameters in the network can be restrictive, while learning the scale parameter is especially beneficial when dealing with inputs at multiple resolutions. 

In this paper, we introduce the Scale-Equivariant UNet (SEUNet), which demonstrates superior generalisation performance on image datasets at different scales when compared with the conventional CNN model and other scale-equivariant models. The main characteristics of out work are as follows: 1) We parameterise convolutional filters with learnable Gaussian derivative filters, instead of using a set of pre-calculated, fixed filter basis. 2) We impose range constraints on learnable scale parameters to ensure coverage of multiple scales, while allowing them to be tuned within disjoint intervals. This frees up model capacity to find an optimal set of scale parameters that adapt to training samples by back-propagation. 

\section{Related Work}
In recent years, group equivariance as an inductive bias for CNNs has influenced the design of several architectures including scale-equivariant convolutional networks. \citet{worrall2019deep} propose deep scale-space (DSS) based on the theory of scale-space and semi-groups to model transformation properties of images under scale transformations, modelling filter rescaling by dilation. However, the DSS is restricted only to integer scale factors, and therefore does not cover a continuous range of scale variations.  To extend DSS to arbitrary scales, \citet{Sosnovik_2021_BMVC,Sosnovik_2021_ICCV} propose Discrete Scale Convolution (DISCO) wherein the equivariance error between the non-integer scale factor with its two nearest integer scale factors is minimised. In Scale-Equivariant Steerable Networks (SESN)  \citep{sosnovik2019scale}, filters are parameterised by trainable linear combination of pre-calculated Hermite basis functions.  These are defined in the continuous scale domain and then projected on pixel grids for a set of given scale factors. Although SESN and DISCO allow the use of arbitrary scale factors, the best set of scale factors are dataset and network dependent and need to be carefully chosen to maximise model performance.

Gaussian scale-space theory \citep{lindeberg1994scale} represents an image as a one-parameter family of gradually smoothed signals, in which the fine scale details are successively suppressed by convolving the image with a set of re-scaled Gaussian filters and Gaussian derivative filters. \cite{lindeberg2022scale} proposes a Gaussian derivative network in which every convolutional filter is constructed as a linear combination of Gaussian derivative filters. The architecture presented in \cite{lindeberg2022scale} is only evaluated on image classification tasks, for which global scale invariance is key to predictive accuracy.  For image segmentation tasks, the output map should scale in proportional to the input, making scale equivariance a necessary property. Similarly, \cite{pintea2021resolution} learn linear combinations of N-th order Gaussian derivative filters to create the  N-Jet convolutional layer. Unlike \citet{lindeberg2022scale} and \citet{sosnovik2019scale} where the scale parameters ($\sigma$) are fixed, the $\sigma$ and sizes of the filters in the N-Jet layer are learned from the data; this frees the network architect from searching and setting scale-related parameters for datasets and networks. However, the $\sigma$ is shared by all filters in a layer, thus limiting the representational capacity of a N-Jet layer.

Our work extends \citet{lindeberg2022scale} from image classification to image segmentation with while also allowing the $\sigma$ of each layer to be learnable similar to \citet{pintea2021resolution}.  Furthermore, we set the scale factors $\sigma$ to lie in disjoint ranges through the layers of the network.

\section{Methodology}
\textbf{Scale transformations and scale equivariance.} 
The scaling operator $S_s$ is defined on a function (image) $f$ thus:
\begin{equation}
    (S_sf)(x)=f(s^{-1}x),\; s>0.
\end{equation}
For $\Phi$ a family of feature mapping operators, scale equivariance means that the scaling transformation should commute with the feature mapping operation according to
\begin{equation}
    \Phi^\prime(S_sf)=S_s(\Phi(f)),
\end{equation}
Where $\Phi^\prime$ denotes some feature map operators within the same family $\Phi$ that operates on the image re-scaled by factor of $s$. We refer to cases with $s>1$ as up-scalings and to cases with $s<1$ as down-scalings.
\subsection{Parameterising convolutional filters, layer by layer}\label{section:Filters}
The 1D Gaussian filter at scale $\sigma$ is written as $G(x;\sigma)=\frac{1}{\sigma\sqrt{2\pi}}e^{-\frac{x^2}{2\sigma ^2}}$ which can be extended to 2D isotropic Gaussian filters as $G(x,y;\sigma)=G(x;\sigma)G(y;\sigma)$. Then the 2D Gaussian derivatives can be defined by the product of the partial derivatives on $x$ and on $y$:
\begin{equation}
    G^{i,j}(x,y;\sigma)=\frac{\partial^{i+j}G(x,y;\sigma)}{\partial x^i \partial y^j}=\frac{\partial^i G(x;\sigma)}{\partial x^i}\frac{\partial^j G(y;\sigma)}{\partial y^j}
\end{equation}
\textbf{Filter construction.} In conventional CNNs, a bank of filters $F^l$ of size $[C_l, C_{l-1}, h, w]$ is used to map an input image $f^0$ or feature map $f^{l-1}\in \mathbb{R}^{C_{l-1}\times H\times W}$ into $f^l\in \mathbb{R}^{C_l\times H\times W}$ by convolution. Here $l\geq 1$ is the layer index, $(H,W)$ and $(h,w)$ denotes the size of the feature map and the size of filter, respectively. 
Padding is applied therefore the size of the feature output remains the same as the input. We compute scale-space feature maps by convolving with groups of filters at different scales, with each convolutional filter a linear combination of Gaussian derivative filters:
\begin{equation}
    F^l_k(c_l, x, y, \sigma^l_k; c_{l-1}) = \sum_{i,j\geq 0}^{i+j\leq N} \alpha^l_{i,j, c_l, c_{l-1}} G^{i,j}(x, y; \sigma_k^l),
\label{equ:filter_def}
\end{equation}
where $\alpha^l_{i,j, c_l, c_{l-1}}\in\mathbb{R}$ are learnable, and independent of $k$ which indexes scale settings within a layer. We describe these next, first for the first layer from image to features, and then how features are composed in subsequent layers.  In detail, the $C_l$ channels in $F^l$ are divided into $\gamma$ groups denoted by $F^l_k\in \mathbb{R}^{\frac{C_l}{\gamma}\times C_{l-1}\times h\times w}$, $k\in\{1,\ldots,\gamma\}$. The first layer maps the input image $f^0$ into $f^1=F^1\star f^0$, $f^1\in\mathbb{R}^{\gamma\times\frac{C_1}{\gamma}\times H\times W}$, by convolving with filters $F^1_k(c_1,x,y,\sigma^1_k; c_0)$ at position $(x,y)$, input channel $c_0\in\{\tt r,g,b\}$ and output channel $c_1\in\{1,\cdots,\frac{C_1}{\gamma}\}$. The first dimension in $f^1$ represents the scale axis, with $\gamma$ scales  indexed by $k$.

Note that the subscript $k$ of $\sigma^l_k$ denotes that the scale parameter varies across groups in the same convolutional layer $l$, for all $l$, but is shared across filters in the same group. The $\alpha^l_{i,j,c_l,c_{l-1}}$ in equation~\eqref{equ:filter_def} does not have a group index $k$ in its subscript, as we share these learnable weights between groups, in order to ensure that the convolution kernels generated in different groups are consistent in shape, and do not mix separated scale factors, thus ensuring scale equivariance. When $\gamma=1$, the $F^l$ degenerates to N-Jet convolutional filter in which the $\sigma$ is shared for the complete layer \citep{pintea2021resolution}. We visualise the constructed multi-scale filters in Appendix~\ref{appd:c}.

\textbf{Scale convolution in hidden layers.} For layers $l\geq 2$ feature maps are divided into $\gamma$ groups $f^{l-1}\in \mathbb{R}^{\gamma\times \frac{C_{l-1}}{\gamma}\times H\times W}$, each representing the response to a specific scale in $f^{l-1}_k\in \mathbb{R}^{\frac{C_{l-1}}{\gamma}\times H\times W}$. We again use equation~\eqref{equ:filter_def} to construct $\gamma$ groups of filters $F^l=[F^l_1,\cdots,F^l_\gamma],\; F^l_k\in \mathbb{R}^{\frac{C_l}{\gamma}\times\frac{C_{l-1}}{\gamma}\times h\times w}$ to convolve with $f_k^l$. After the first layer, we define the network architecture to have $f^l_k = F^l_k\star f^{l-1}_k$: in subsequent layers each group of filters acts only on a subset of scale-matched channels. This is in contrast to the first layer ($l=1$) where the filters act on the entire image to generate $f^1_k:=(f^1)_k = F^1_k\star f^0$, with all colour channels contributing to the $\gamma$ scale-specific channels in $f^1$ indexed by $k$.   We thus have the learnable coefficients $\alpha$ and $\sigma_k^l$ range over channel indices
\begin{equation}
    \{\alpha^l_{i,j, c_l, c_{l-1}} | c_0 \in \{{\tt r, g, b}\}, c_l = 1,\ldots, \frac{C_l}{\gamma} \}\mbox{ and }
    \{\sigma^l_k | k=1,\ldots,\gamma \}.
\label{equ:params}
\end{equation}


Thus, the propagation of information captured by composition of layer-wise convolutions does not mix information from different scales.  The restriction on network connections to scale-matched layer outputs is designed to maintain equivariance to input rescaling at layer outputs under composition. Although acting $F^l_k$ on the entire $f^l$ can be another option, an attempt in \cite{sosnovik2019scale} shows that introducing inter-scale interaction also introduces extra equivariance error, and leads to lower performance.  We further tune the successive scale factors $\sigma^l_k$ to track the increase in the receptive field with depth.  

\subsection{Imposing range constraints on $\sigma_k^l$}\label{section:range_cons}
The trainable parameters in equation~\eqref{equ:params} in the filters include the scale parameters $\sigma^l_k$ learned during back-propagation. However, leaving them to be tuned completely freely may lead to a problem: all $\sigma^l_k$s in the same layer may have the same value, which means constructed filters are redundant, limiting the scale diversity of filters. As our original intention is that the network can achieve scale equivariance by learning multi-scale convolutional filters, we introduce the following constraints to separate $\sigma_k^l$ values to lie in disjoint intervals.
\begin{equation}
    \label{equ:range}
    \sigma^l_k(x) = \frac{a^l_k-b^l_k}{2}\tanh{x}+\frac{a^l_k+b^l_k}{2},\;a^l_k>b^l_k,\; b\geq0 
\end{equation}
where $a^l_k$ and $b^l_k$ are hyper-parameters for the upper and lower bounds for $\sigma$ of filters at the $l^{th}$ layer and the $k^{th}$ group. $x$ is a trainable real variable.  By setting an appropriate set of $a^l_k$ and $b^l_k$: multi-scale filters can be constructed as per equations~\eqref{equ:filter_def}. Once $\sigma^l_k$ is known, the following formula used in \cite{pintea2021resolution} is employed to determine the size $\tau^l_k$ of filters $f_k^l$: 
\begin{equation}
    \label{equ:size}
    \tau^l_k=2\left \lceil 2\sigma_k^l \right \rceil +1,\; \sigma_k^l>0
\end{equation}
This enables us to train the size of the receptive field. In the encoder path of the UNet (layer 1-8), we gradually increase $\sigma$, to increase receptive field size. This is in keeping with \citet{lindeberg2022scale}. In the decoder path (layer 11-18), we gradually decrease $\sigma$. Layer 9-10 are the bottleneck layers. An ablation study with regard to the setting of $\sigma_k^l$ can be found in Appendix~\ref{app:ablation} which demonstrates the benefit of imposing range constraints on $\sigma_k^l$.

\subsection{Parallelising training by simultaneous optimisation of multiple loss functions}
\begin{figure}[!h]
    \centering
    \includegraphics[scale=0.5]{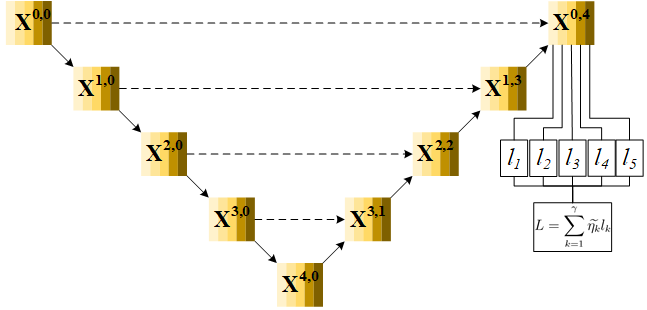}
\caption{The architecture of the Scale-Equivariant UNet (SEUNet) model proposed here. Each square node in the graph represents a convolution block that consists of two convolutional layers. In each square node, five rectangles in different colours denotes five groups of filters that are parameterised by different $\sigma_k^l$s, but share the same $\alpha$. All filters with the same colour form an independent sub-network, and all sub-networks have their own prediction and loss functions $\{l_k\;|\;k\in\{1,\cdots,5\}\}$. All sub-networks can be trained simultaneously in an end-to-end fashion by minimising the combined loss function.}
    \label{fig:architecture}
\end{figure}

As described in section~\ref{section:Filters}, filters in each layer are divided into $\gamma$ groups and each group of filters operate only on an non-overlapping subset of feature maps with no inter-scale feature interactions. Thus, features learned by these $\gamma$ groups of filters can be trained in a mutually independent fashion. A penultimate convolution layer for feature fusion creates a score that is passed to a softmax function, followed by the calculation of loss function. This, however, mixes multi-scale information and destroys the scale equivariance of the features. Therefore, to train all groups of filters simultaneously while maintaining equivariance between multi-scale features, we propose to minimise a weighted combination of multiple loss functions, with each of them acting only on a single group of filters. In detail, given the ground truth $y$ and feature map $f^{L-1}_k$ that is produced by filters $F^{L-1}_k$ in a $L$-layer network, a 1$\times$1 convolution with softmax activation is used to map $f^{L-1}_k$ into a probability map $\hat{y}_k$ for each of $C$ classes for every pixel in the $H\times W$ image.  The loss function used to train $F^{L-1}_k$ is the norm cross-entropy loss:
\begin{equation}
    \label{equ:loss}
    l_k(y,\hat{y}_k)=-y{\rm log}(\hat{y}_k),\;y\in\{0,1\}^{C\times H\times W},\hat{y}_k\in\mathbb{R}^{C\times H\times W}, \; k=1,\ldots, \gamma.
\end{equation}
The overall loss function is defined as:
\begin{equation}
    \label{equ:loss}
    L=\sum_{k=1}^\gamma \widetilde{\eta}_k l_k(y,\hat{y}_k),\;\widetilde{\eta}_k= \frac{\eta_k+\frac{1}{\gamma}}{\sum_{k=1}^{\gamma}{(\eta_k+\frac{1}{\gamma})}},\;\mbox{ where } 0\leq  \eta_k\leq 1,\;\sum_{k=1}^\gamma \eta_k = 1,\;\sum_{k=1}^\gamma \widetilde{\eta}_k=1.
\end{equation}
$\eta_k$ is a weighting factor that assigned to $F_k$ to characterise the relative importance between scales. It is quite plausible that the loss function is minimised by a dominating contribution from a specific scale indexed by $k$, driving all other $\eta_k$ to zero, a phenomenon called competitive exclusion. It is to maintain some contribution from features acquired at multiple scales that we introduce the additive constant $(1/\gamma)$ in $\widetilde{\eta}_k$. This constrains the trainable $\widetilde{\eta}_k$ to be in the range $[\frac{1}{2\gamma}, \frac{\gamma+1}{2\gamma}]$. In practice, we initialise $\eta_k$ to $\frac{1}{\gamma}$ and use the \textit{softmax} function to normalise $\eta_k$ to guarantee $\sum_{k=1}^\gamma \eta_k$ = 1. Figure~\ref{fig:architecture} shows the entire structure of the model proposed here.
\subsection{Final Prediction Generation}
The SEUNet generates probability maps $\hat{y}_{k,n}$, $k=1,\ldots,\gamma$ from learned filters with different scales/sizes for each pixel $n$. For each pixel $n$, let $\hat{y}_{k, n, c}$ be the probability of predicting class $c\in C$ by classifier indexed by scale $k$. Given an image with unknown scale information and these $\gamma$ probability maps, we explore the following strategies to generate the final segmentation map.

\textbf{Arithmetic mean ensemble.} For each pixel $n$ the final segmentation map is obtained from $\arg\max_c(1/\gamma)\sum_k \hat{y}_{k,n,c}$. 

\textbf{Per-pixel prediction selection based on prediction confidence.} Let $(k, n, c^*) = \arg\max_c \hat{y}_{k,n,c}$ and $(k, n, c') = \arg\max_{c\neq c^*} \hat{y}_{k,n,c}$.  Then $\delta_{n, k}:=(\hat{y}_{k, n, c^*} - \hat{y}_{k, n, c'})$, the difference between the largest and second-largest class probability is a measure of the predictive confidence of classifier $k$. We choose the most \textit{confident} prediction ($k^* = \arg\max_k \delta_{n, k}$, so $c^* = \arg\max_c \hat{y}_{k^*, n, c}$) for pixel $n$ as its final predicted label. We denote this strategy P\_Dist.


\textbf{Per-pixel prediction ensemble based on prediction confidence.} To mitigate against a concern of an incorrect prediction made with high confidence, we propose P\_Ens, a per-Pixel ensemble strategy that weights multiple predictions based on their confidence.  Thus multiple less confident predictions can compensate in test cases where the highest confident prediction may be incorrect. 

The detailed process of generating final prediction using P\_Dist or P\_Ens strategies is described in Appendix~\ref{app:algorithm}.
\section{Experiments and Results}
\subsection{Datasets}
\textbf{MoNuSeg dataset}
The MoNuSeg dataset \citep{kumar2019multi} is a multi-organ nucleus segmentation dataset. The training set includes 37 images of size 1000$\times$1000 from 4 different organs (lung, prostate, kidney, and breast). The test set contains 14 images with more than 7000 nucleus boundary annotations. All images are scaned at 40$\times$ magnification. A 400$\times$400 window slides through the images with a stride of 200 pixels to separate each image into 16 tiles for training and testing.\\
\textbf{BCSS dataset} The Breast Cancer Semantic Segmentation (BCSS) dataset \citep{amgad2019structured} consists of 151 H\&E stained whole-slide images and ground truth masks corresponding to 151 histologically confirmed breast cancer cases. Tissue types of the BCSS dataset consists of 5 classes (\romannumeral1)tumour, (\romannumeral2)stroma, (\romannumeral3)inflammatory infiltration, (\romannumeral4)necrosis and (\romannumeral5)others. We set aside slides from 7 institutions to create our test set and used the remaining images for training. Shift and crop data augmentation, random horizontal and vertical flip were adopted to enrich training samples. Finally, 3154 and 1222 pixel tiles of size 512$\times$512 were cropped for training and testing, respectively.

\subsection{Degree of equivariance}
To quantitatively compare the degree to which our proposed method preserves scale equivariance relative to other scale-equivariant convolutional layers, we rescale  $N$ test images $f_i\mapsto S_s(f_i)$ by scale factor $s$, extract feature maps $\Phi(\cdot)$ and $\Phi^\prime(\cdot)$, and then calculate the equivariance error:
\begin{equation}
   \Delta_s = \frac{1}{N}\sum_{i=1}^N \frac{\|S_s\Phi(f_i)-\Phi^\prime( S_s(f_i))\|_2^2}{\|S_s\Phi(f_i)\|_2^2}.
\label{equ:equivariance}
\end{equation}
where $\Phi(\cdot)$ and $\Phi^\prime(\cdot)$ denotes a sequence of convolutional operations with filters parameterised by different sets of $\sigma_k$.
\subsection{Compared methods}
We use the UNet architecture as a backbone and replace the conventional convolutional layers with different types of scale-equivariant convolution to generate 3 scale-equivariant UNet variants (SESN: the UNet with SESN layers; DISCO: the UNet with DISCO layers; SEUNet: the UNet with the proposed Gaussian derivative layers, the model generates $\gamma$ segmentation maps). For the UNet with conventional convolutional layers, the number of filters at each depth are 60, 120, 240, 480, 960. For a fair comparison, all of UNet variants have the same number of scales (refers to the hyper-parameter $\gamma$). for the SESN and DISCO model, we set $\gamma$ as 5 and scale factors as $\{1,2,3,4,5\}$, therefore the size of filters at each scale is $\{3,5,7,9,11\}$. For SESN model, we set the highest order of Hermite polynomial as 4 since it demonstrates the best performance. For the proposed models, we carefully set the lower and the upper bound of $\sigma^l_k$ to set the size of filters (derived from equation~\eqref{equ:size}) of the first layer to be consistent with that of SESN and DISCO. The $\sigma$ range of each layer is shown in Figure~\ref{fig:sigma_cons_BCSS} and ~\ref{fig:sigma_cons_kumar} (black dashed lines). We set the highest order of the Gaussian derivative to be 1, since using higher order derivatives fails to provide better performance. The colour normalisation method proposed in \cite{vahadane2016structure} is used to remove stain colour variation, before training. All models are trained on images at the original scale, scale augmentation is not used in all of our experiments.

We implement the conventional UNet model and our proposed methods. The officially released source code of SESN and DISCO layers is used in our experiments. All Models are implemented in Pytorch \cite{paszke2019pytorch} and trained on one NVIDIA RTX 8000 GPU using the Adam optimiser \cite{kingma2014adam} with weight decay of 10$^{-4}$ to minimise the cross-entropy loss. The training epoch is set as 70, and the initial learning rate for the Adam optimiser is set as 0.015 and then changed according to the 1cycle learning rate policy \cite{smith2019super}. The batch size is 20 for training models.
\subsection{Results and Discussion}
In this section, we report the overall segmentation performance of the three UNet architectures followed by ablation studies to analyse the performance gain of our approach. For the BCSS dataset, we use the mean Intersection over Union (mIoU) to measure segmentation performance of models, while for the binary task in the MoNuSeg dataset, we report the IoU score of the nuclei class. Examples of images, masks and segmentation maps generated by models can be seen in Appendix~\ref{app:visualisation}.

\textbf{Evaluation regime.} Since we aim to evaluate models' scale equivariance property, we re-scale the test set by a series of scale factors between 0.25 and 4, with a relative scale ratio of $\sqrt[4]{2}$ between adjacent testing scales.

\begin{table}[t]
\begin{tabular}{l|c|cccccccccc}
\toprule
\multicolumn{2}{c}{Test Scale}        & 0.25           & 0.3            & 0.35           & 0.42           & 0.5            & 0.59           & 0.71           & 0.84           & 1              \\\midrule
\multirow{5}{*}{\makecell[c]{Pred\\Head}} & 1 & 30.25          & 34.04          & 37.12          & \textbf{43.99} & \textbf{52.49} & \textbf{57.71} & 59.87          & 58.98          & 57.17          \\
     & 2 & \textbf{30.39} & \textbf{34.39} & \textbf{37.68} & 43.75          & 49.78          & 56.32          & \textbf{60.25} & \textbf{59.94} & \textbf{58.28} \\
     & 3 & 26.62          & 30.76          & 34.06          & 40.36          & 46.62          & 53.58          & 59.02          & 59.78          & 58.20  \\
     & 4 & 22.56          & 26.27          & 29.21          & 34.47          & 40.58          & 49.14          & 56.56          & 58.97          & 57.82 \\
     & 5 & 21.81          & 24.55          & 27.10          & 31.73          & 36.85          & 44.16          & 53.27          & 57.89          & 58.13 \\\midrule\toprule
\multicolumn{2}{c}{Test Scale}          & 1.19           & 1.41           & 1.68           & 2              & 2.38           & 2.83           & 3.36           & 4              & mean           \\\midrule
\multirow{5}{*}{\makecell[c]{Pred\\Head}} & 1 & 51.90          & 43.62          & 34.72          & 27.41          & 23.66          & 21.23          & 19.89          & 19.13          & 39.60          \\
     & 2 & 54.08          & 46.32          & 37.86          & 30.50          & 25.42          & 22.28          & 20.76          & 19.97          & 40.47          \\
     & 3 & \textbf{54.95} & 47.82          & 39.62          & 32.96          & 27.99          & 23.64          & 21.76          & 20.93          & 39.92          \\
     & 4 & 54.70          & \textbf{48.94} & \textbf{41.68} & 35.70          & 31.63          & 27.66          & 24.16          & 22.33          & 38.96          \\
     & 5 & 54.62          & 49.11          & 41.31          & \textbf{35.87} & \textbf{32.26} & \textbf{29.25} & \textbf{26.63} & \textbf{24.64} & 38.19\\\bottomrule 
\end{tabular}
\caption{The mIoU score on the BCSS dataset (highest in bold). Head 1-5 denote 5 classifiers appended to the 5 groups of filters of increasing $\sigma$ at the last layer. The actual $\sigma$ values are shown in Figure~\ref{fig:sigma_cons_BCSS}. The last column is the mean mIoU score over all scales.}
\label{table:bcss}
\end{table}
\textbf{Scale specific predictions for SEUNet}. Table~\ref{table:bcss} summarises the mIoU score of the proposed method on re-scaled test images. Although we offer 3 strategies to generate the final segmentation maps from $\gamma$ predictions, we first check the performance of each group of filters. As seen in the table, the best prediction head shifts to the one with larger $\sigma$ values, as the scale of images increase (although prediction head 2 gives best prediction for scales between 0.25-0.35, the performance gap to the prediction head 1 is very small). This is consistent with our intuition that to capture the same information from enlarged images, filter sizes should also increase. 

\begin{table}[t]
\centering
     \begin{tabular}{ll|ccc|ccc}
     \toprule
             Dataset & Metric & CNN   & SESN  & DISCO & Arithm & P\_Dist & P\_Ens \\\midrule
        \multirow{2}{*}{MoNuSeg} & IoU   & 54.94 & 57.29 & 59.77 & 59.96  & 59.93  & \textbf{59.98}\\
                    & E\_Err & 0.64 & 0.57 & 0.39 & \multicolumn{3}{c}{\textbf{0.26}} \\\bottomrule
        \multirow{3}{*}{BCSS} & mIoU   & 34.48 & 35.36 & 40.51 & 40.51  & 42.10  & \textbf{42.43}  \\
        & E\_Err & 0.74 & 0.72 & 0.61 & \multicolumn{3}{c}{\textbf{0.54}} \\\bottomrule
\end{tabular}
\caption{Experimental results of different methods on MoNuSeg and BCSS dataset. E\_Err denotes the equivariance error of feature maps generated by the last layer (before the final prediction generation step).}
\label{tab:models}
\end{table}

\begin{figure*}[t]
\begin{center}
    \subfigure[BCSS dataset]{
        \label{fig:bcss_curve}
        \includegraphics[width=0.47\linewidth]{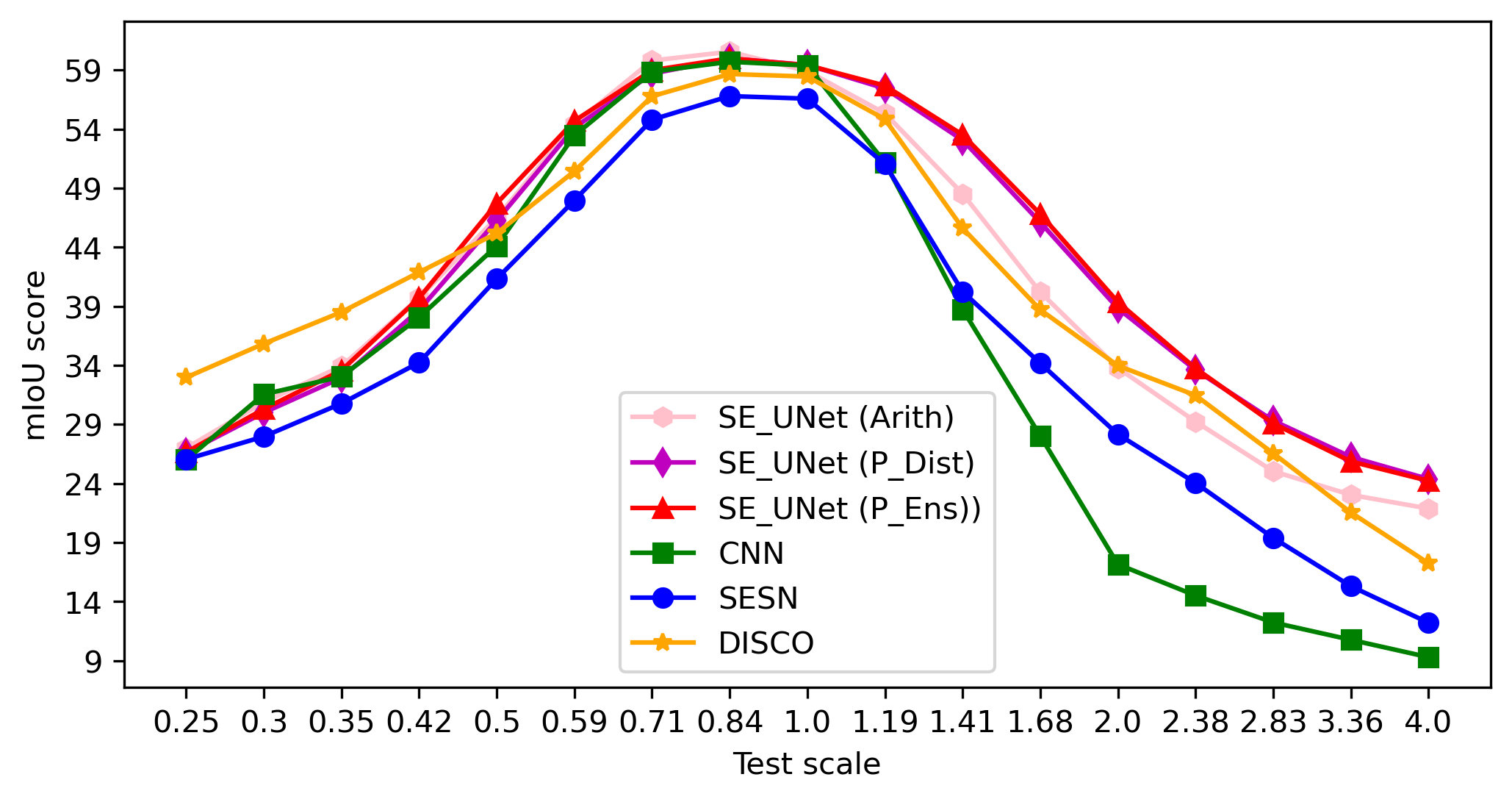}
    }
    \subfigure[MoNuSeg dataset]{
        \label{fig:kumar_curve}
        \includegraphics[width=0.47\linewidth]{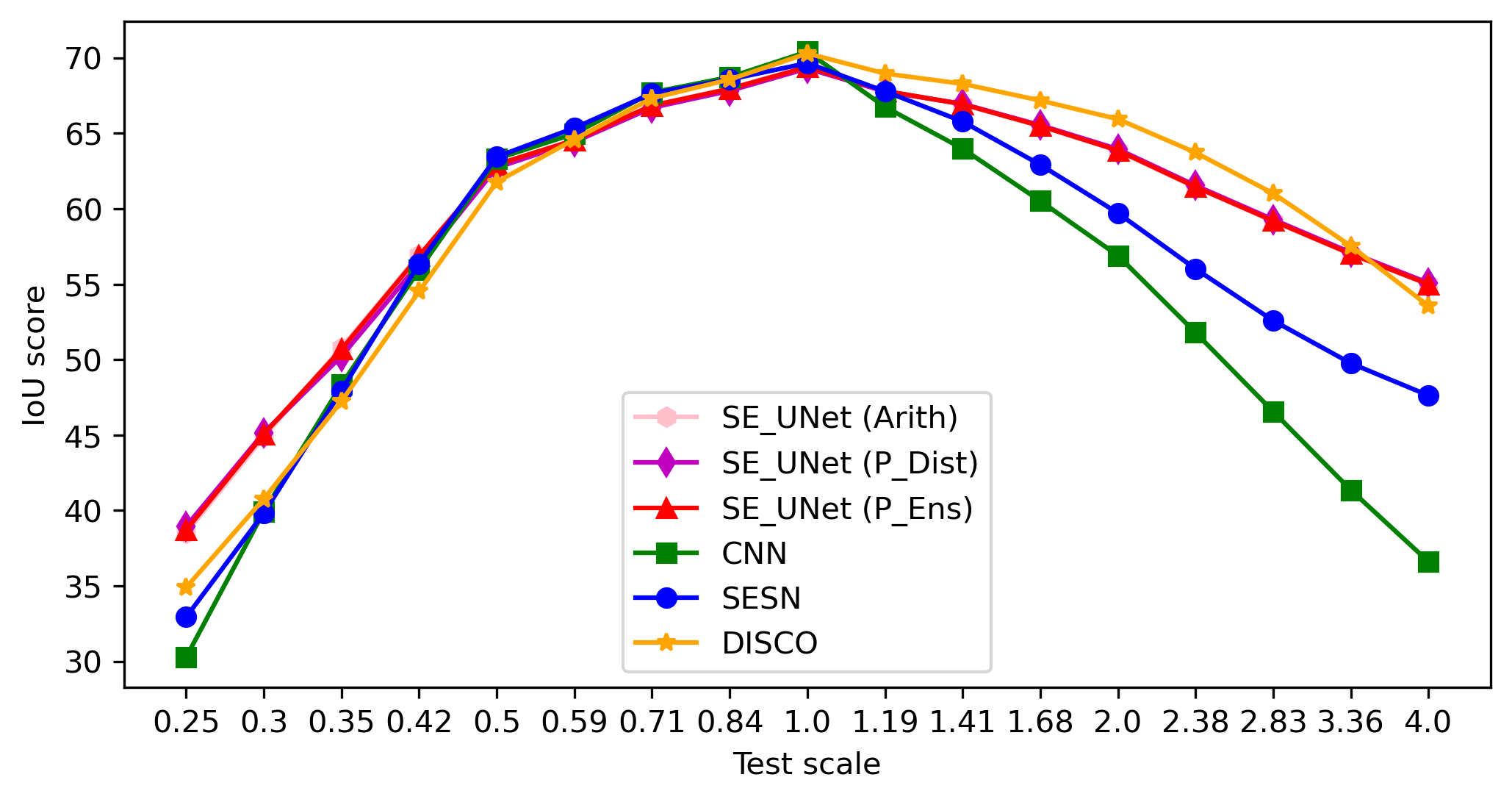}
    }
\end{center}
   \caption{Comparison of the Per-scale mIoU score between models.}
\label{fig:onecol}
\end{figure*}

\textbf{Comparison with CNN baseline and other equivariant methods.} Figure~\ref{fig:bcss_curve} and~\ref{fig:kumar_curve} show the per-scale test performance of different approaches on the BCSS and the MoNuSeg datasets. The performance of all models are similar when the test scale is close to the original training scale. However, as the test scale moves away from the training scale, the performance of the conventional CNN drops significantly. Although the performance of SESN, DISCO and the proposed method also drop as the test scale changes, these models look more robust than the CNN. Table~\ref{tab:models} compares models in terms of the averaged mIoU scores over different scales, the equivariance of feature map at the last layer, and the number of trainable parameters. As observed from the table, the proposed SEUNet with different prediction ensemble/selection strategies outperforms all compared methods, particularly on the BCSS dataset, while using fewer parameters. In terms of prediction strategy, simply averaging the prediction demonstrates the worst performance (on the BCSS dataset). This suggests that mixing features of all scales equally without considering the possibility that scale-specific filters have different contributions to the prediction is not the optimal choice. The proposed P\_Dist strategy surpasses the arithmetic mean ensemble by 1.59 points on the BCSS dataset. Moreover, the P\_Ens further boosts the performance by 0.33 points, when compared with the P\_Dist. This comparison validates the effectiveness of the P\_Dist and P\_Ens strategy. We report the equivariance error $\Delta_s$ in Table~\ref{tab:models} computed using the final layer outputs of all three networks on both datasets. We note that lower equivariance error correlates with better segmentation performance when tested on multi-scale images.
\section{Ablation Study}\label{app:ablation}
In section~\ref{section:range_cons}, we propose to constrain the value of $\sigma^l_k$ in some non-overlapping ranges, to ensure that the constructed filters capture relevant patterns at different scales. Here, to verify the effectiveness of imposing range constraints and to trace the origin of the performance gain of the proposed method, we conduct the following two ablation experiments.

\textbf{1) Fix $\sigma^l_k$ values.} Instead of constraining $\sigma^l_k$ to the range of $(a^l_k,\;b^l_k)$ in equation~\eqref{equ:range}, we fix the $\sigma^l_k$ to be $\frac{a^l_k+b^l_k}{2}$.

\textbf{2) $\sigma^l_k$ being trained freely.}
\begin{figure*}[!t]
    \centering
    \subfigure[$\sigma^l_k$ is constrained (BCSS).]{
        \includegraphics[width=0.45\linewidth]{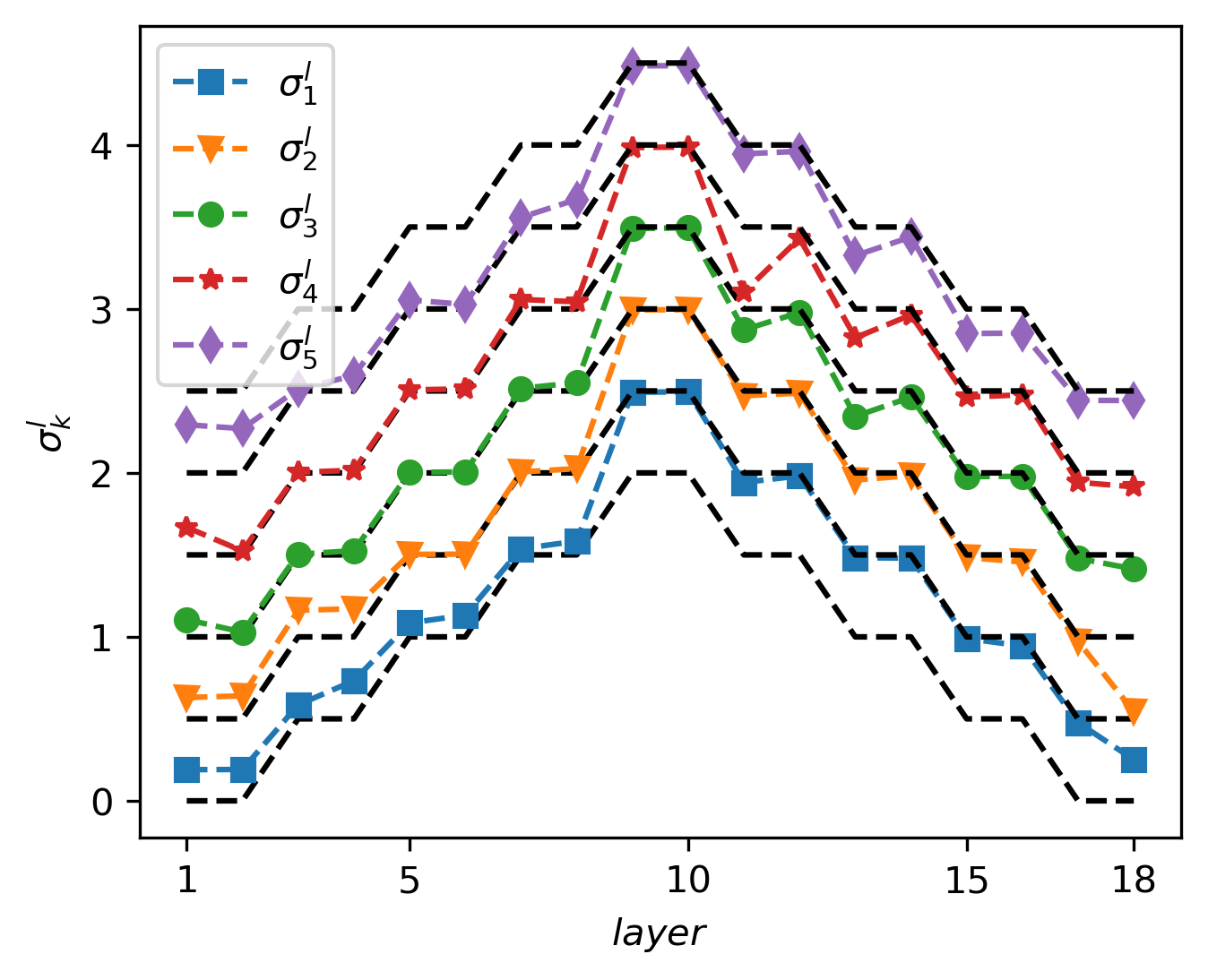}
        \label{fig:sigma_cons_BCSS}
    }
    \subfigure[$\sigma^l_k$ is constrained (MoNuSeg).]{
        \includegraphics[width=0.45\linewidth]{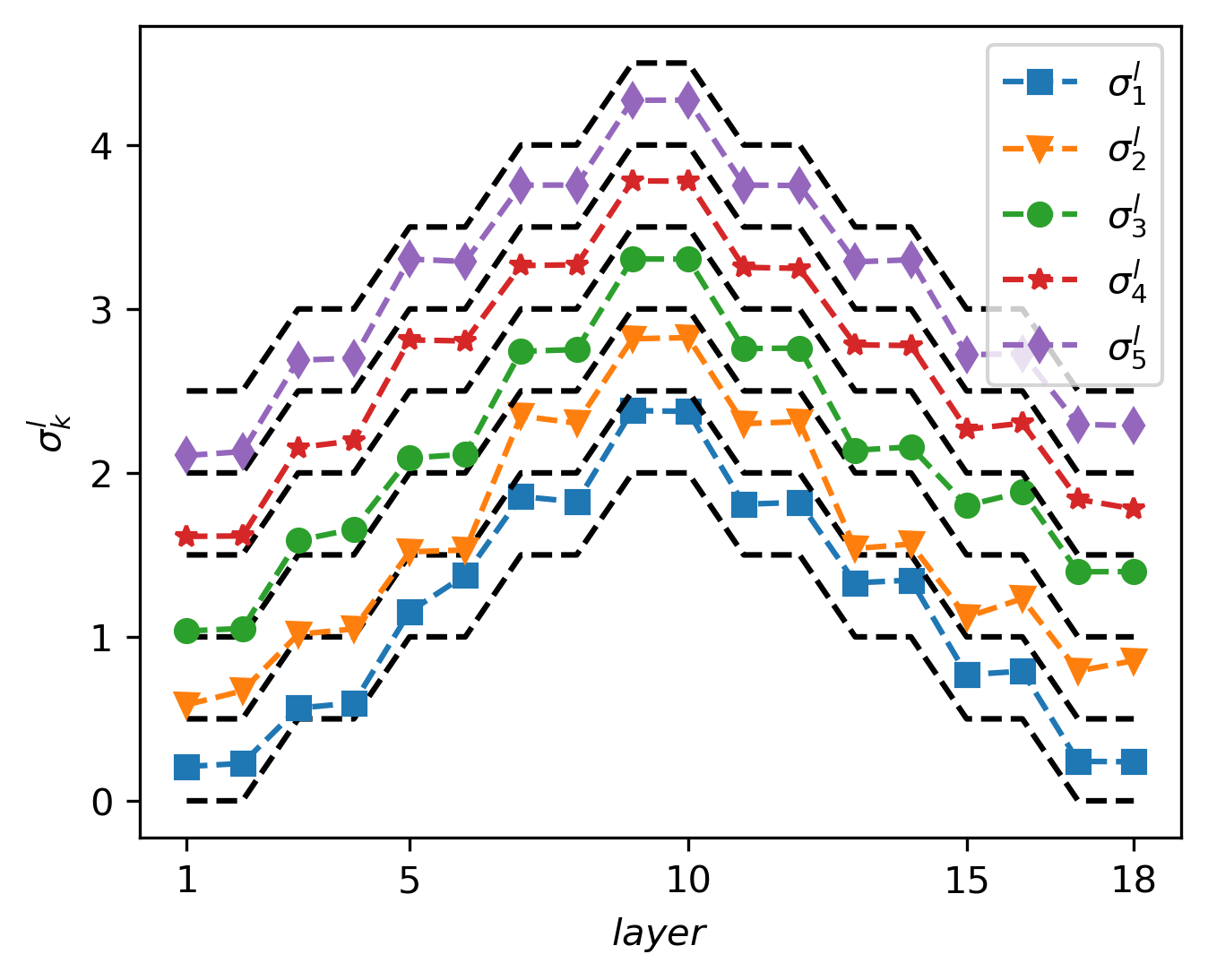}
        \label{fig:sigma_cons_kumar}
    }\\
    \subfigure[$\sigma^l_k$ is trained freely (BCSS).]{
        \includegraphics[width=0.45\linewidth]{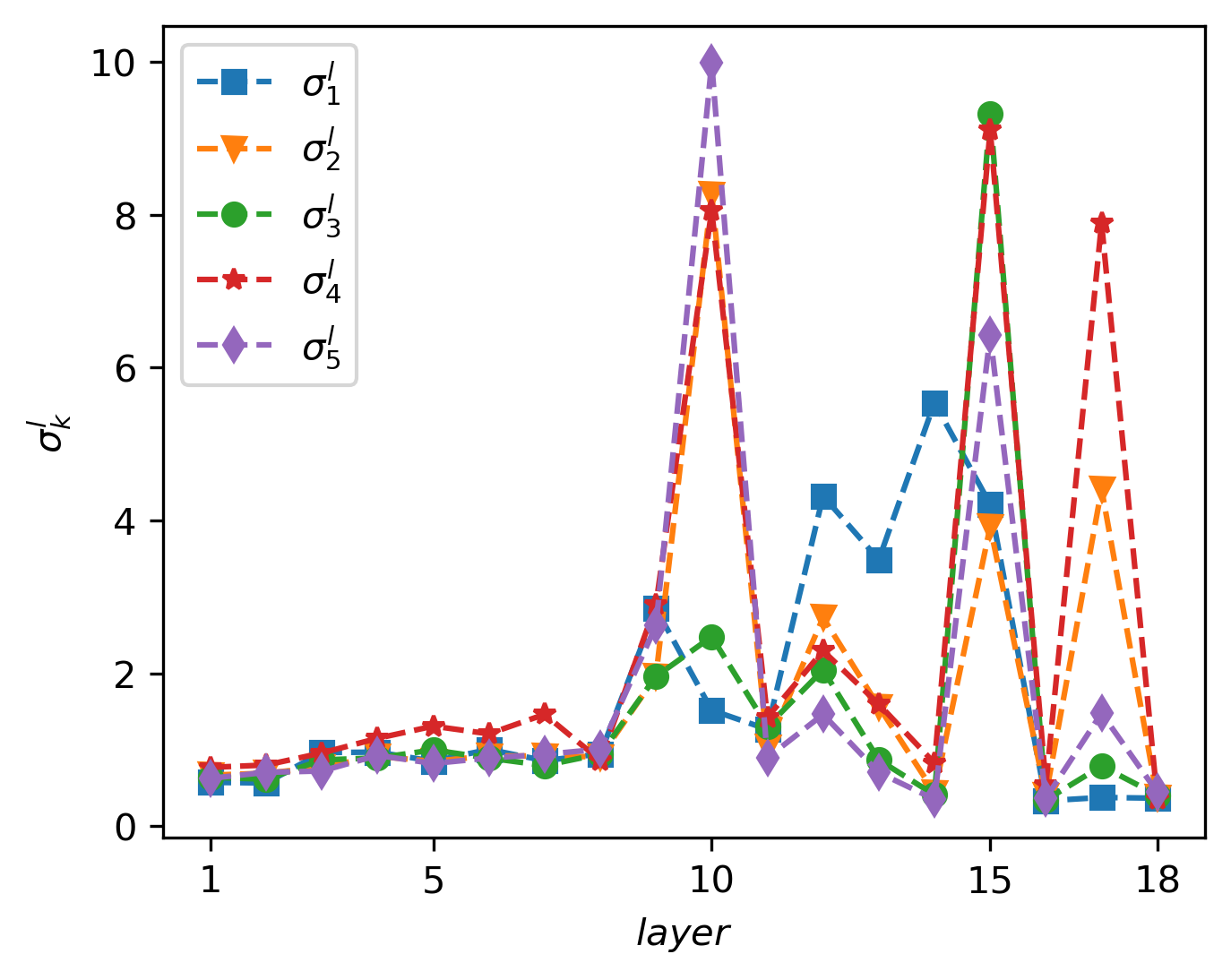}
        \label{fig:sigma_free_BCSS}
    }
    \subfigure[$\sigma^l_k$ is trained freely (MoNuSeg).]{
        \includegraphics[width=0.45\linewidth]{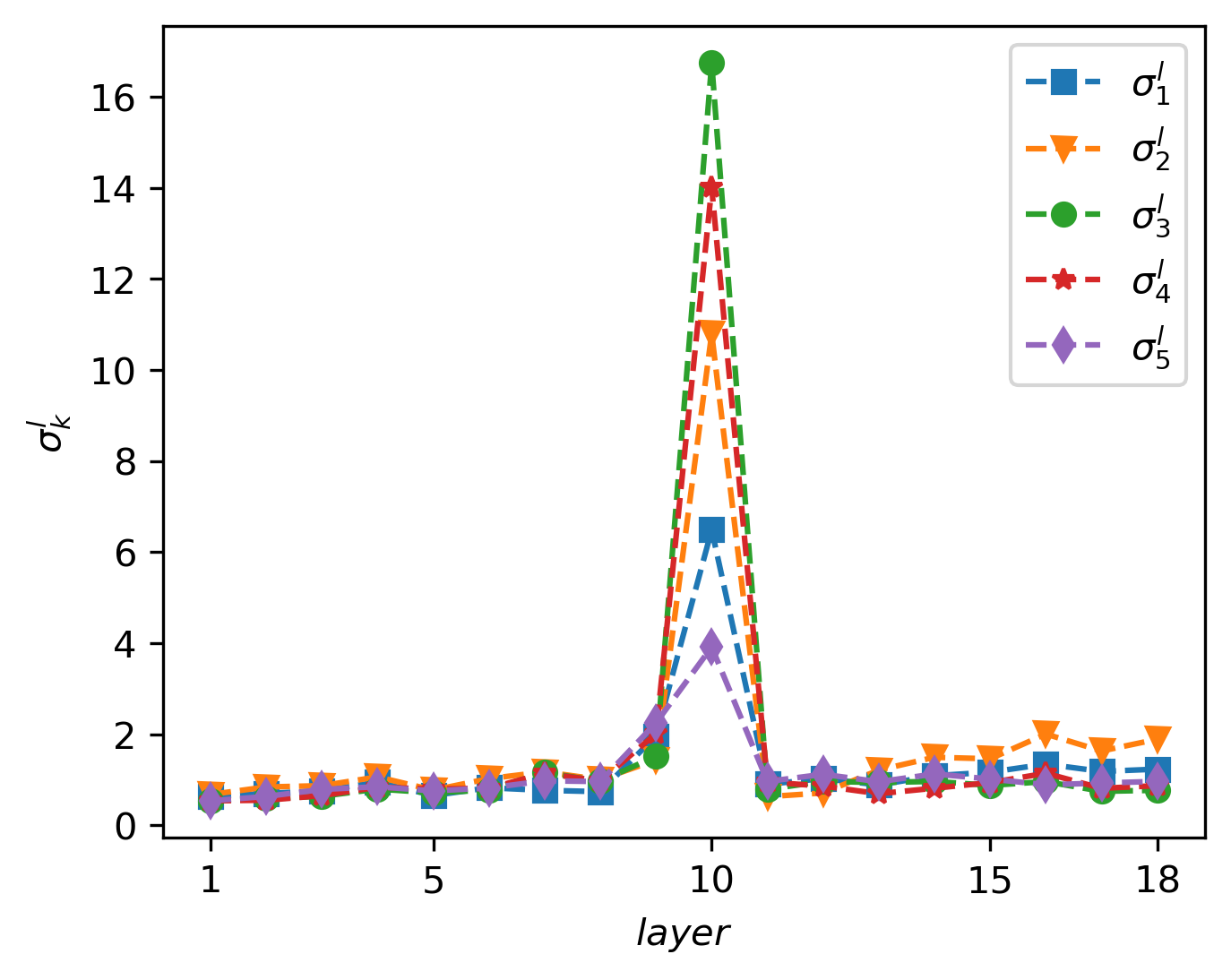}
        \label{fig:sigma_free_kumar}
    }
    \caption{The $\sigma^l_k$ values of filters in different layers. (a) and (b) The $\sigma^l_k$ of models trained on the BCSS and the MoNuSeg datasets under range constraints. (c) and (d) The $\sigma^l_k$ of models trained on the BCSS and the MoNuSeg datasets without imposing range constraints. The black dashed lines in (a) and (b) show the range of $\sigma^l_k$ of layers.}
    \label{fig:sigma_layers}
\end{figure*}
\begin{table}[!b]
    \centering
    \begin{tabular}{c|ccc|ccc}
        \toprule
         & \multicolumn{3}{c}{MoNuSeg} & \multicolumn{3}{c}{BCSS}  \\
         & Cons & Fixed & Free & Cons & Fixed & Free \\\midrule
        mIoU & \textbf{59.98} & 58.36 & 57.40 & \textbf{42.43} & 41.45 & 39.02\\\bottomrule
    \end{tabular}
    \caption{mIoU of SEUNets trained in different $\sigma$ settings.}
    \label{tab:ablation}
\end{table}

Table~\ref{tab:ablation} summarises the performance achieved by models trained under different $\sigma^l_k$ settings. The mIoU score reported in the table is the mean of per-scale mIoU score obtained by P\_Ens. As shown in the table, models trained with range constraint outperform models with $\sigma^l_k$ being fixed or trained completely freely, on both datasets. Figure~\ref{fig:sigma_layers} shows the $\sigma^l_k$ values of models trained under different settings. As can be seen in Figure~\ref{fig:sigma_free_BCSS} and~\ref{fig:sigma_free_kumar}, allowing $\sigma^l_k$ to be trained freely results in the case that multiple $\sigma^l_k$s converge to the same value. This is detrimental to the feature representation of the model since the same $\sigma^l_k$ means that the same scale of the generated filters, thus the resultant feature map is also the same (because the coefficient $\alpha$ is shared between filters in different groups). Therefore, features are redundant and are not scale equivariant. This is the reason why the model trained without range constraint demonstrates the worst performance. The model trained with fixed $\sigma^l_k$ values performs slightly better than the freely trained one, since $\sigma^l_k$s are non-overlapping, multi-scale filters can be constructed to extract information from different scaled images. However, the manually selected $\sigma^l_k$s may not be the optimal choice that fits the dataset best. Moreover, the optimal set of $\sigma^l_k$s may vary from dataset to dataset. This motivated our choice in equation~\eqref{equ:range} to train $\sigma^l_k$s to remain in disjoint intervals. As observed from Figure~\ref{fig:sigma_cons_BCSS} and ~\ref{fig:sigma_cons_kumar}, $\sigma^l_k$s trained under constraint deviate from fixed values. And also, the learned $\sigma^l_k$s are quite different on the BCSS and the MoNuSeg datasets. Another benefit of imposing range constraints on $\sigma^l_k$ is to reduce the computational complexity of the model. In our experiments, we observe that the model trained with range constraints required only $\frac{1}{3}$ the training time of the freely trained one. Because the filter of smaller size is less computationally intensive when performing convolution operations. For example, in Figure~\ref{fig:sigma_cons_BCSS} and~\ref{fig:sigma_cons_kumar}, the maximum $\sigma^l_k$ values are 4.5 and 9.99 (layer 10), respectively, and the corresponding filter sizes are 19 and 41 (calculated from equation~\eqref{equ:size}). Therefore, the amount of computation required by the latter is $\approx4.65\times$ that of the former when convolving with an image.
\section{Conclusion}
In this paper, we propose a Scale-Equivariant UNet (SEUNet) to address the challenge of generalising neural network segmentation on histopathology images to unseen scales. Firstly, we parameterise multi-scale filters by linearly combining groups of Gaussian derivative filters. The constructed filters are then used to learn scale-space representations that have a built-in scale-equivariant property. We constrain filter scales to be both trainable yet cover disjoint ranges. This is useful for finding dataset-adapted scale parameters. The extensive experimental results on two public datasets demonstrate that the proposed SEUNet achieves state-of-the-art performance. However, since we learn the Gaussian derivatives during training, these derivatives should be updated after each round of $\sigma^l_k$ updating, which is more computationally expensive than using a pre-calculated filter basis. In the future, more range constraints will be explored to enable improved performance.
\section*{Acknowledgement}
The authors acknowledge the use of the IRIDIS High Performance Computing Facility, and associated support services at the University of Southampton, in the completion of this work. The first author (Yilong Yang) is supported by China Scholarship Council under Grant No. 201906310150.

\bibliography{yang22}
\let\cleardoublepage\clearpage
\appendix
\section{Pseudo-Code of Generating Final Prediction}\label{app:algorithm}
We first calculate the prediction confidence, the difference between the largest and the second largest probability, then the weighting factor $w_k$ of prediction $\hat{y}_k$ is determined by the \textit{softmax} function, giving larger weights to confident predictions.
\begin{algorithm2e}[!h]
  \caption{Per-pixel Prediction Selection/Ensemble based on Probability Distance.}
  \label{alg:P_Ens}
  \LinesNumbered 
  \KwIn{Probability vectors $[\hat{y}_{1},\cdots,\hat{y}_{\gamma}]$, $\hat{y}_{k}\in\mathbb{R}^C$, $k\in\{1,\cdots,\gamma\}$; $C$ is the number of classes. $Strategy\in \{P\_Dist$ or $P\_Ens\}$.}
  $D=[\;]$ \tcp*{Recording distance.}
  \For{$k=1,2,...,\gamma$}{
        $p_{max}$ = max($\hat{y}_{k}$)\\
        $p_{max\_idx}$ = $\arg\max(\hat{y}_{k}$)\\
        $\hat{y}_{k,p_{max\_idx}}=-1$\\
        $p_{second\_max}$ = max($\hat{y}_{k}$)\\
        $D.{\rm append}(p_{max}-p_{second\_max})$\\
        $\hat{y}_{k,p_{max\_idx}}=p_{max}$\\
    }
    \If{Strategy==P\_Dist}{
            $k=\arg\max D$\\
            $\hat{y}=\underset{C}{\arg\max}(\hat{y}_k)$\\
    }
    \If{Strategy==P\_Ens}{
        \For{$k=1,2,...,\gamma$}{
            $w_k=\frac{e^{D_k}}{\sum_{k=1}^\gamma e^{D_k}}$\\
        }
        $\hat{y}=\underset{C}{\arg\max}(w_k\hat{y}_k)$\\
    }
    \KwOut{$\hat{y}$}
\end{algorithm2e}
\section{Visualisation of Multi-Scale Filters}\label{appd:c}
\begin{figure*}[!h]
    \centering
    \subfigure[$\sigma_1=0.5$]{
        \includegraphics[scale=0.26]{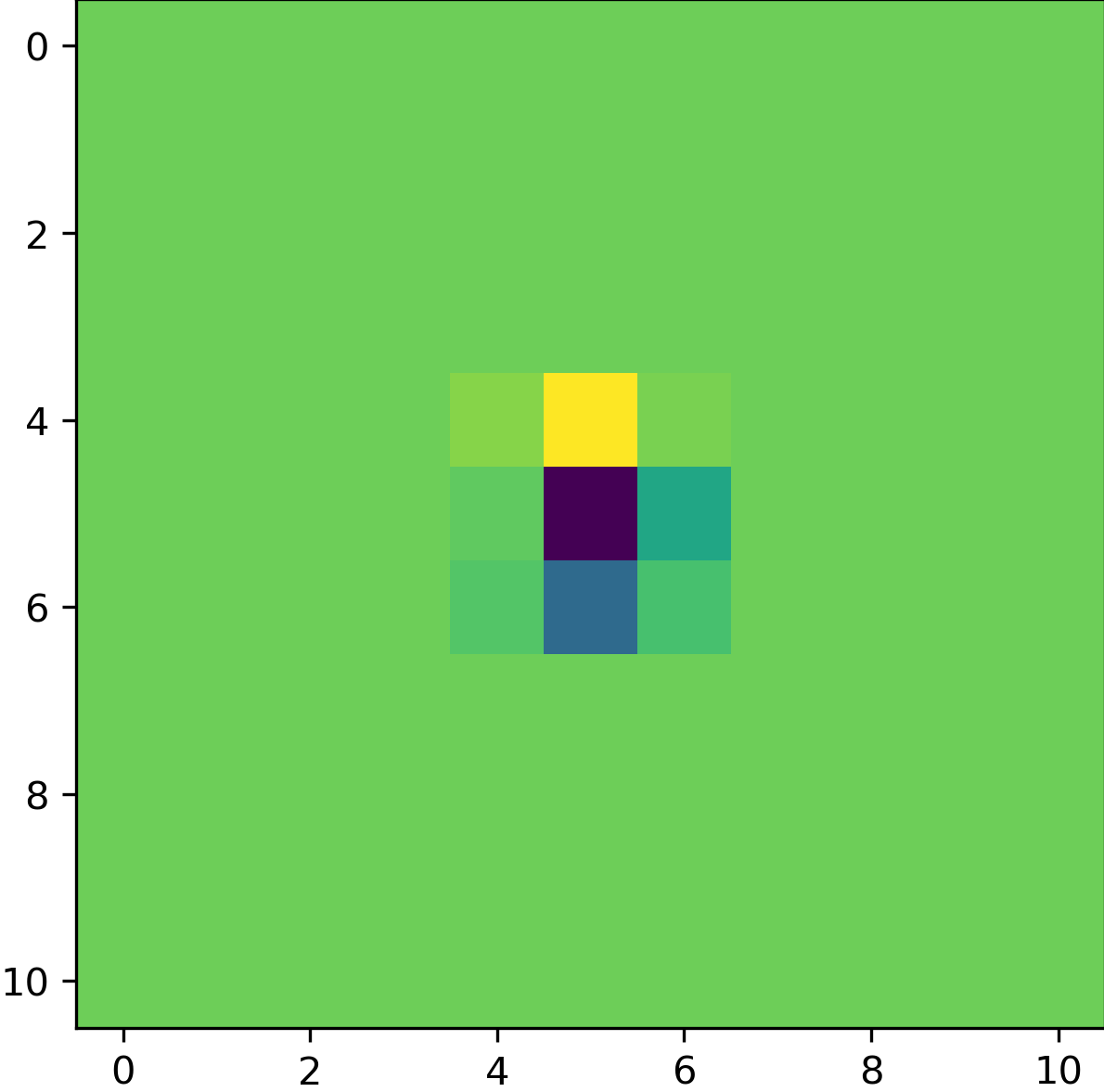}
    }
    \subfigure[$\sigma_2=1$]{
        \includegraphics[scale=0.26]{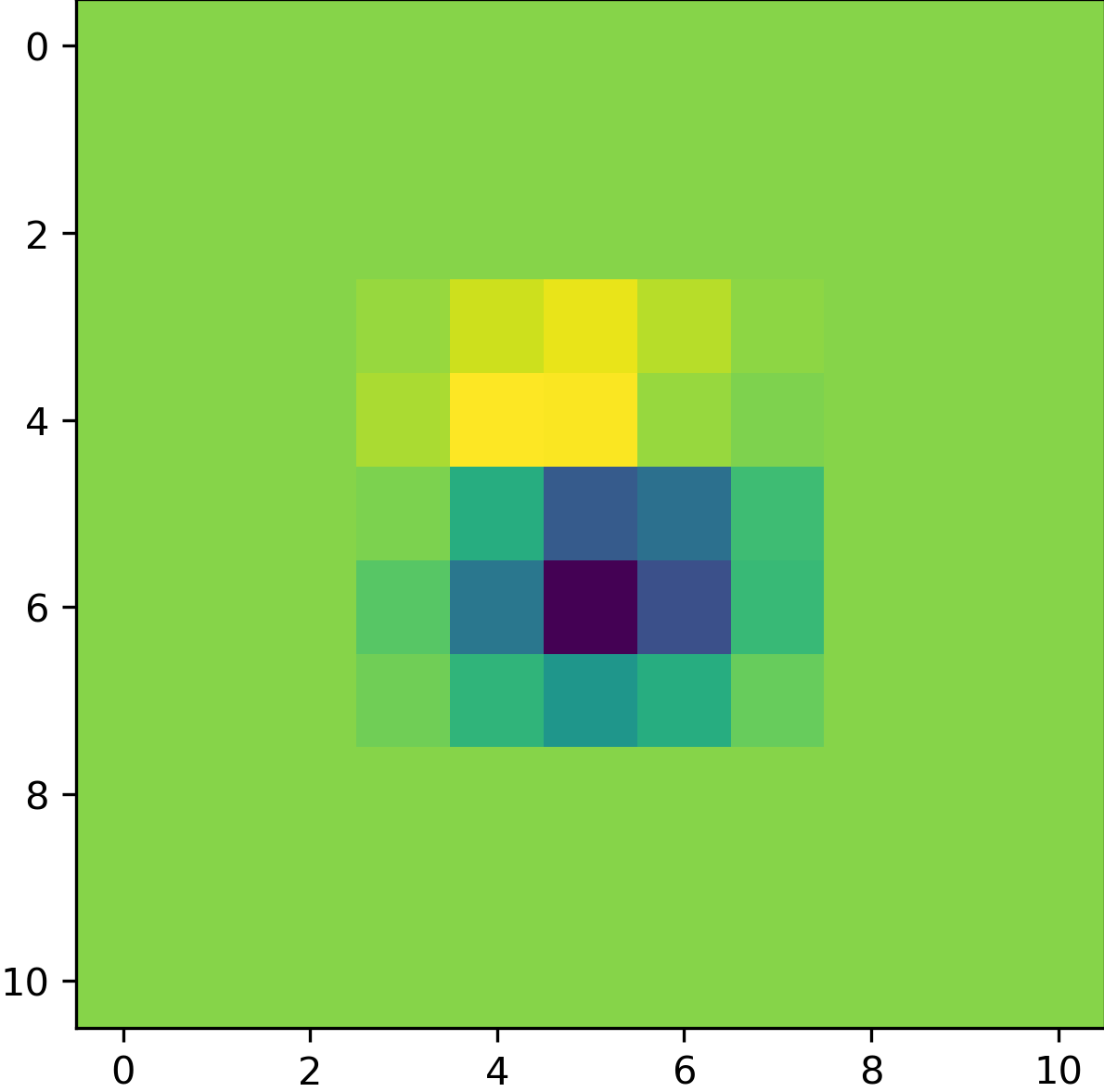}
    }
    \subfigure[$\sigma_3=1.5$]{
        \includegraphics[scale=0.26]{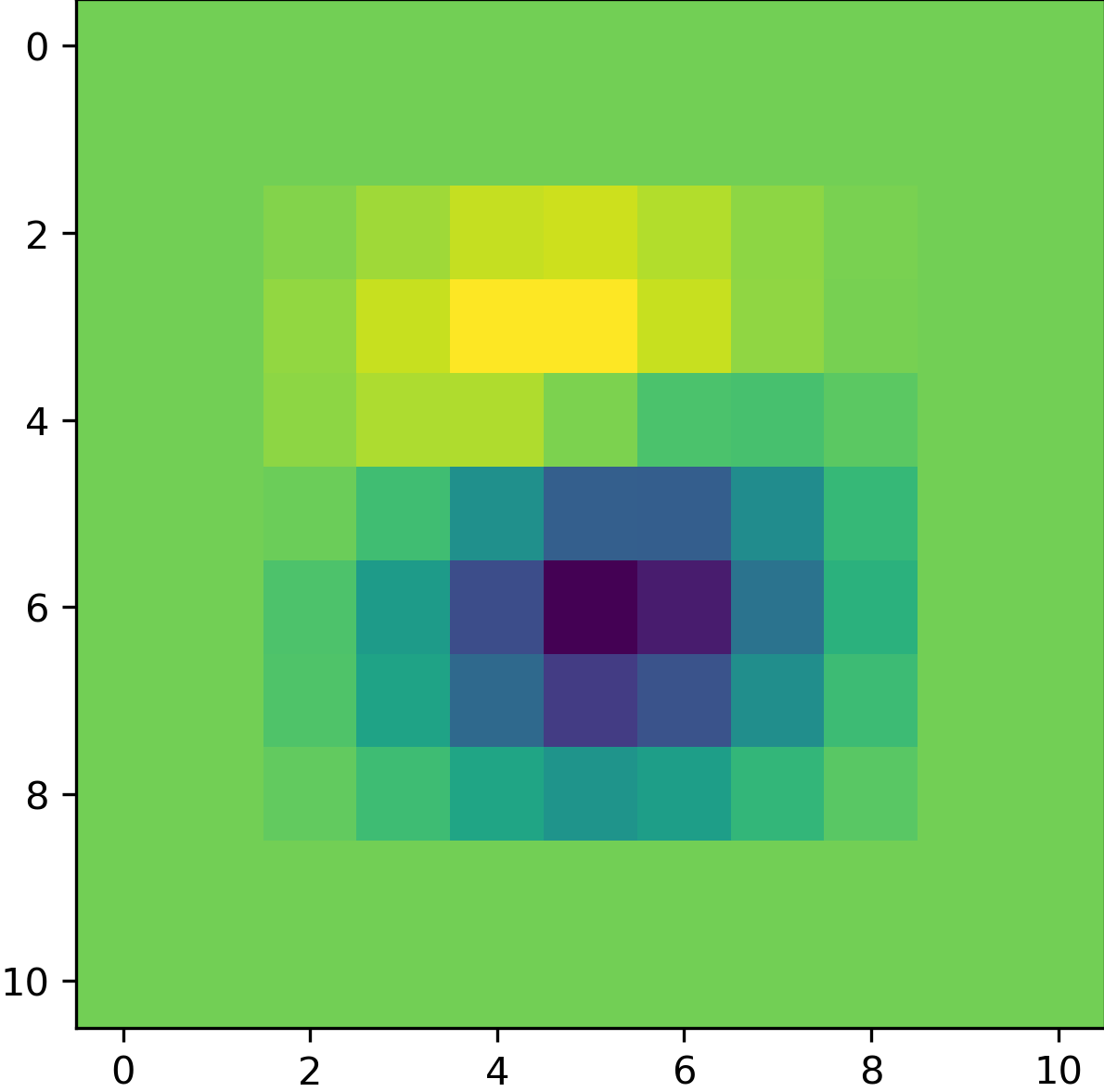}
    }
    \subfigure[$\sigma_4=2$]{
        \includegraphics[scale=0.26]{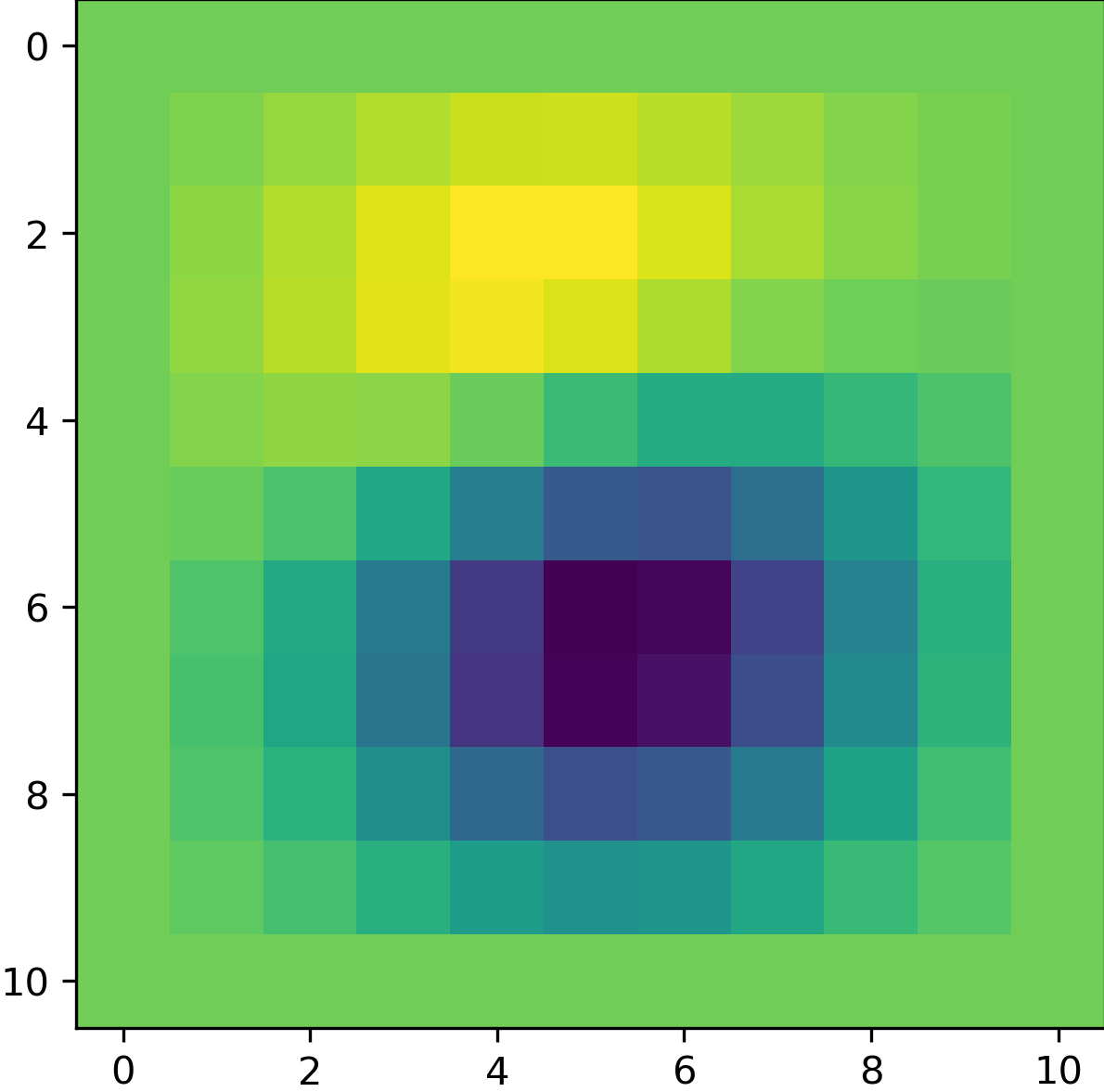}
    }
    \subfigure[$\sigma_5=2.5$]{
        \includegraphics[scale=0.26]{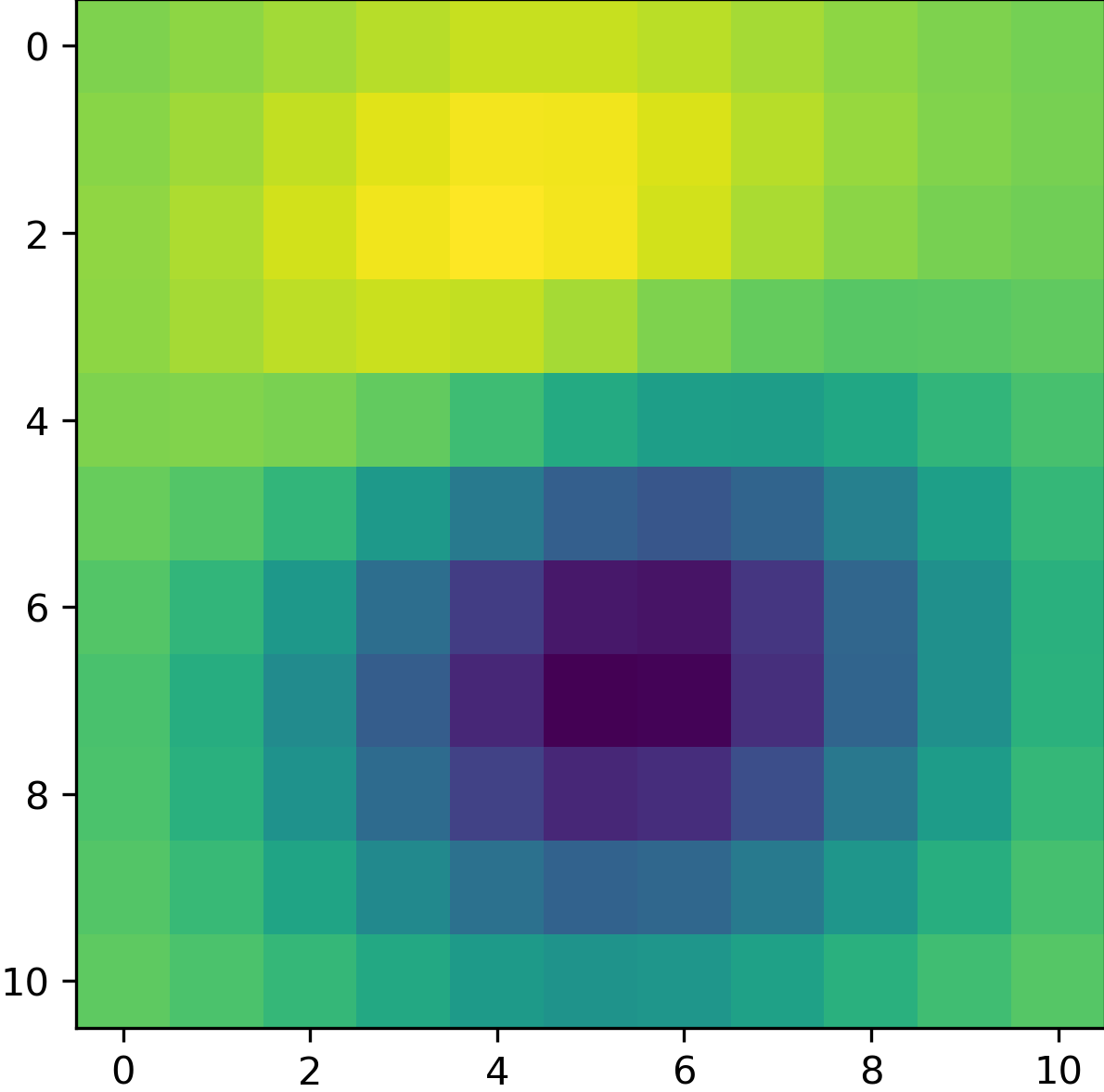}
    }
    \caption{Constructed multi-scale filters. From left to right, the effective size of the filter gradually enlarges as the $\sigma$ increases.}
    \label{fig:filters}
\end{figure*}

Given a set of scale parameters $\{\sigma_1,\cdots,\sigma_\gamma\}$ and the number of scales $\gamma$, the filter of each scale can be constructed by:
\begin{equation}
\begin{aligned}
    F_k(x,y;\sigma_k)=\sum^{i+j\leq N}_{0\leq i,\; 0\leq j}\alpha_{i,j}\frac{\partial^{i+j}}{\partial x^i\partial y^j}G(x,y;\sigma_k),\;k\in\{1,\cdots,\gamma\}
\end{aligned}
\label{equ:vis_first_layer}
\end{equation}
Here we visualise the multi-scale filters generated with a set of predefined $\sigma$ and randomly initialised coefficients ($\alpha$). As shown in Figure~\ref{fig:filters}, filters are similar in shape but vary in scale.
\section{Visualising the Equivariance Error}
\begin{figure*}[!t]
    \centering
    \subfigure[s=0.25]{
        \includegraphics[scale=0.35]{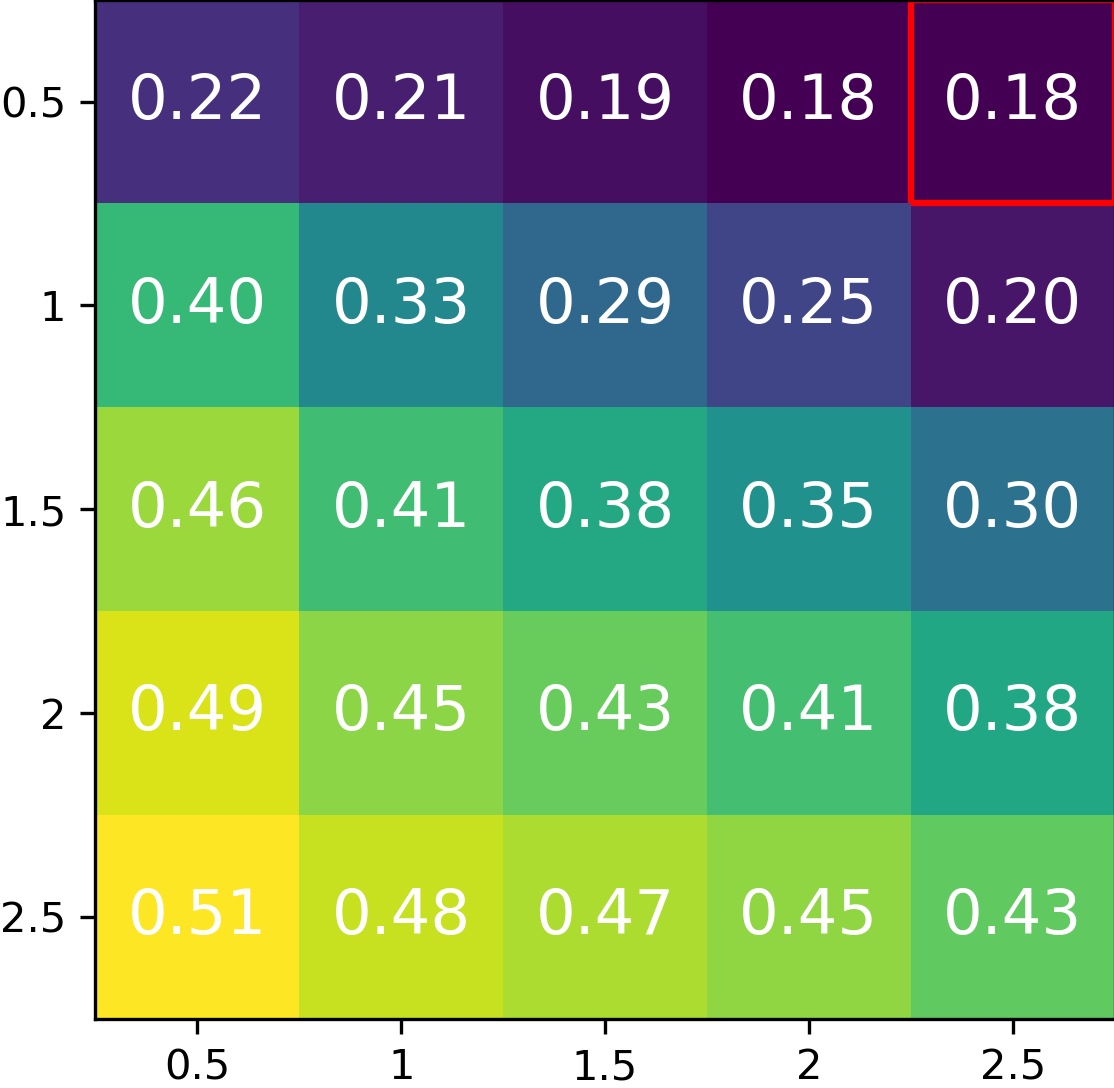}
    }
    \subfigure[s=0.3]{
        \includegraphics[scale=0.35]{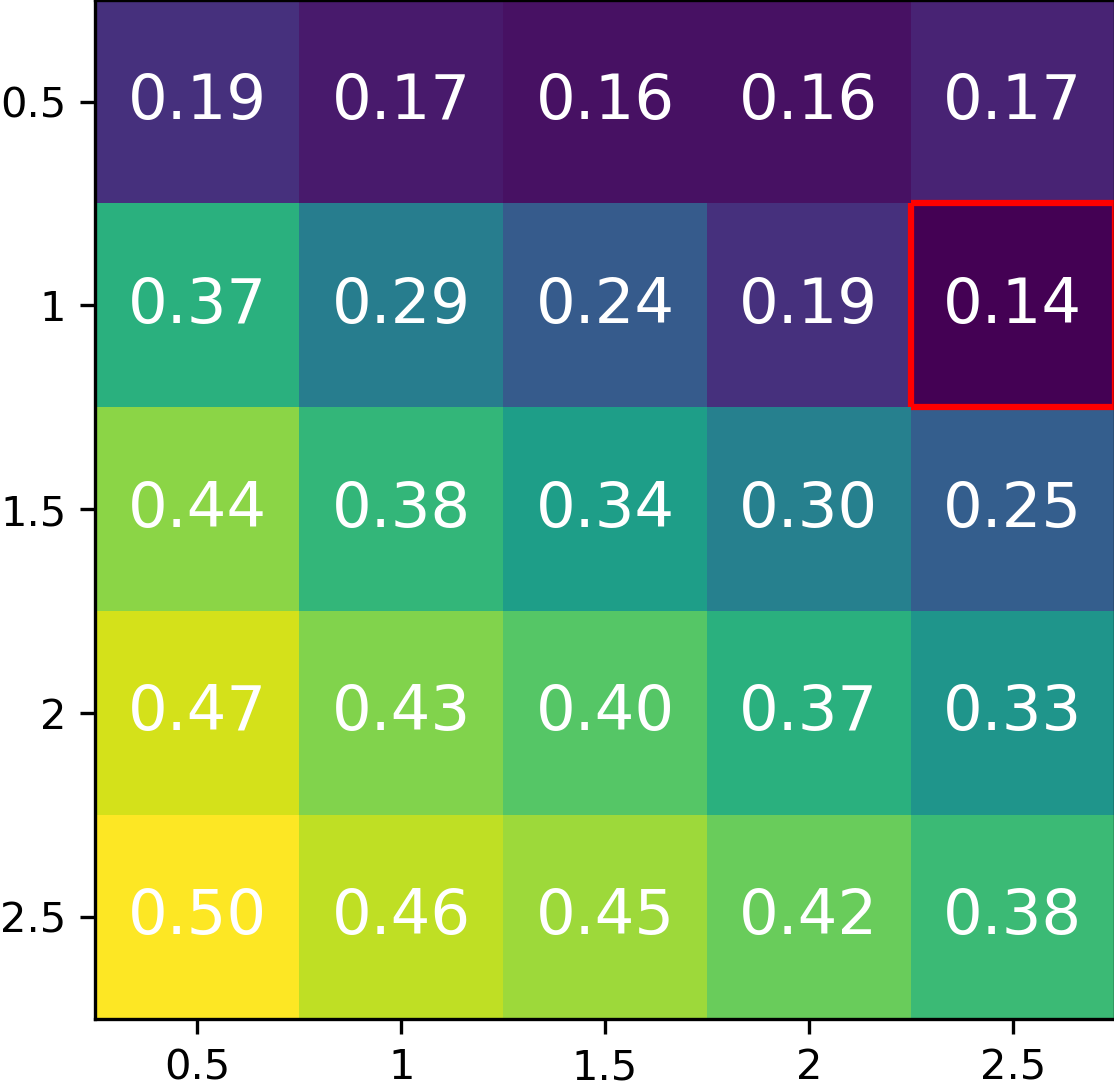}
    }
    \subfigure[s=0.35]{
        \includegraphics[scale=0.35]{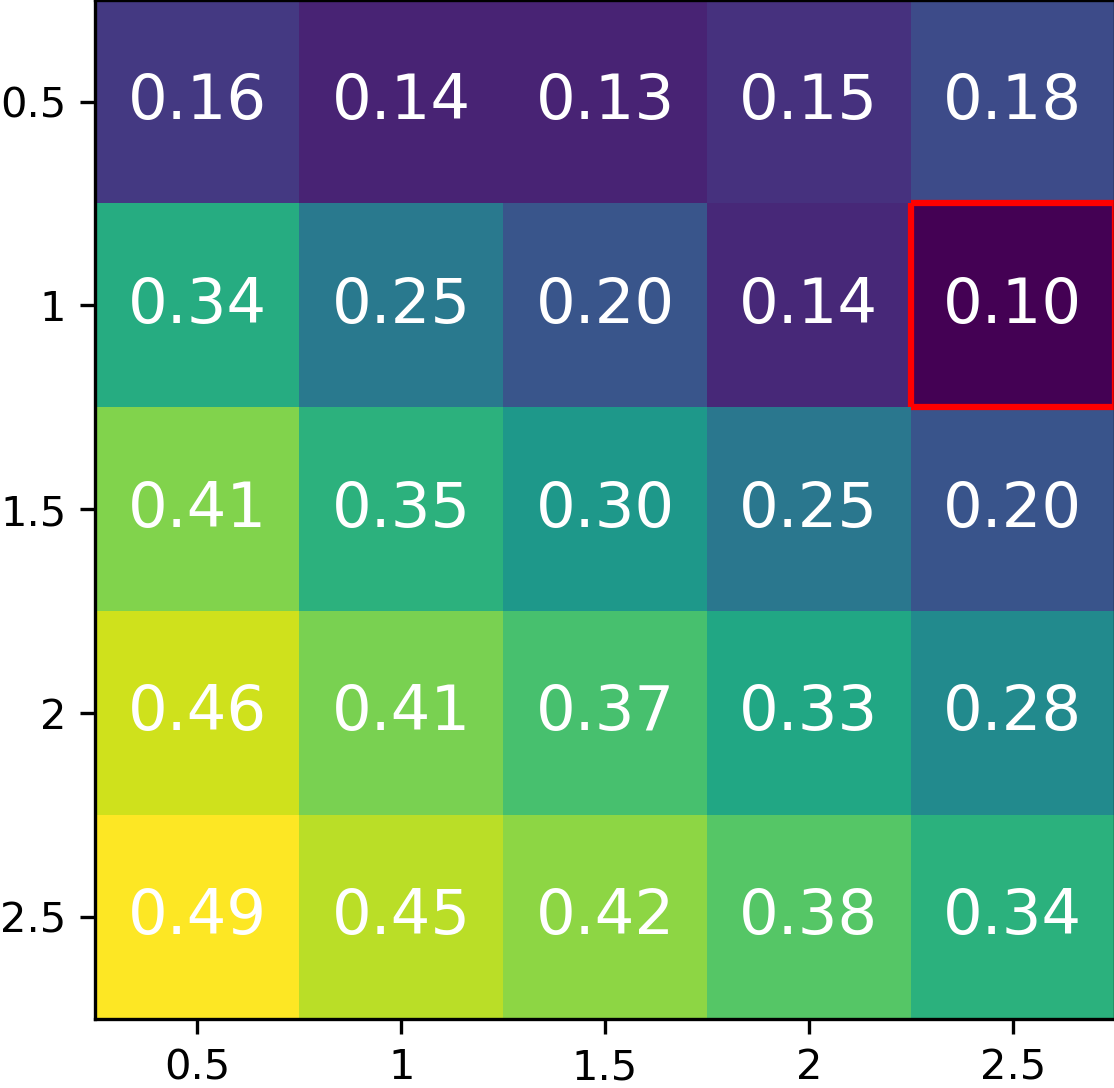}
    }
    \subfigure[s=0.42]{
        \includegraphics[scale=0.35]{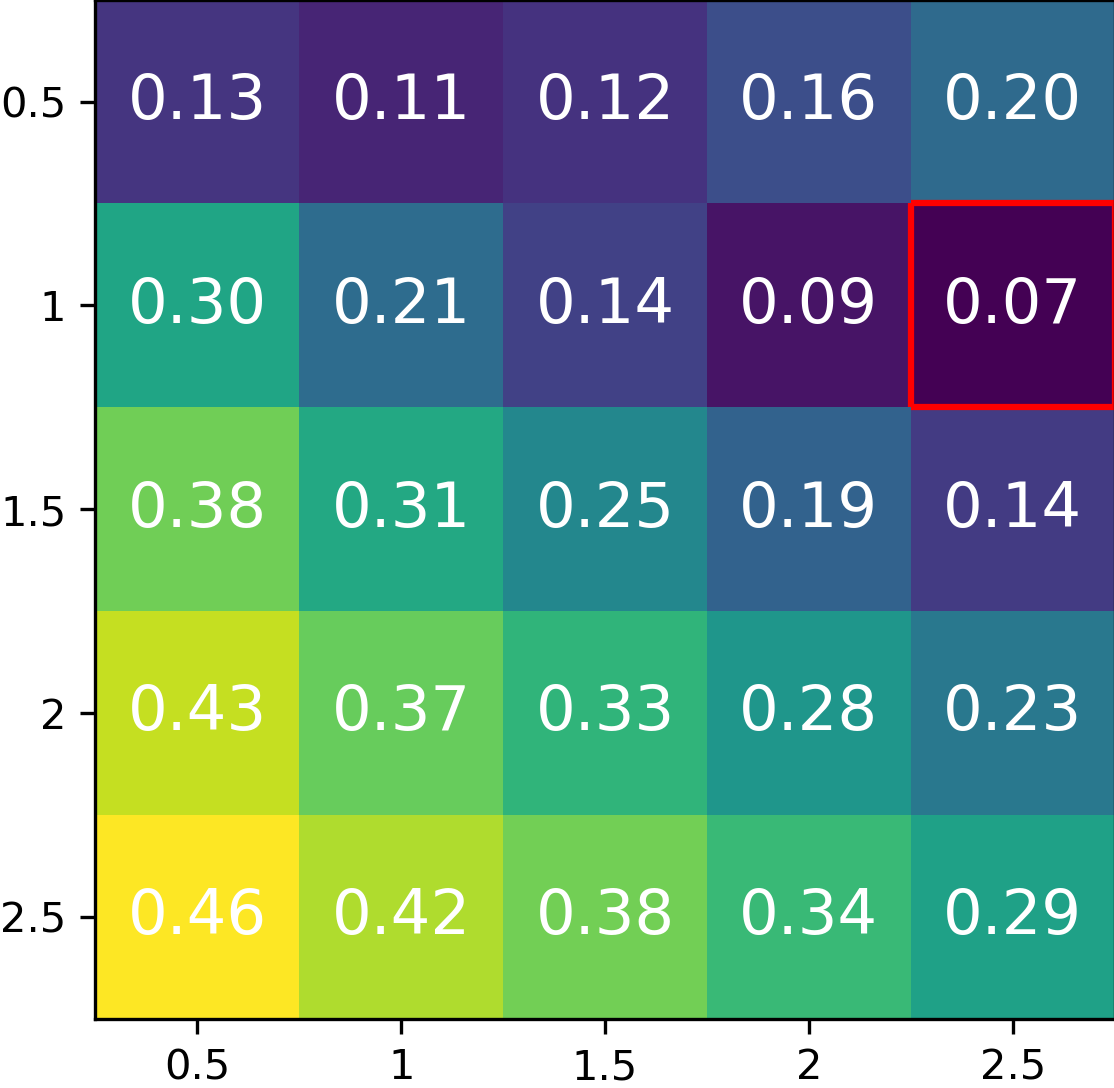}
    }\\
    \subfigure[s=0.5]{
        \includegraphics[scale=0.35]{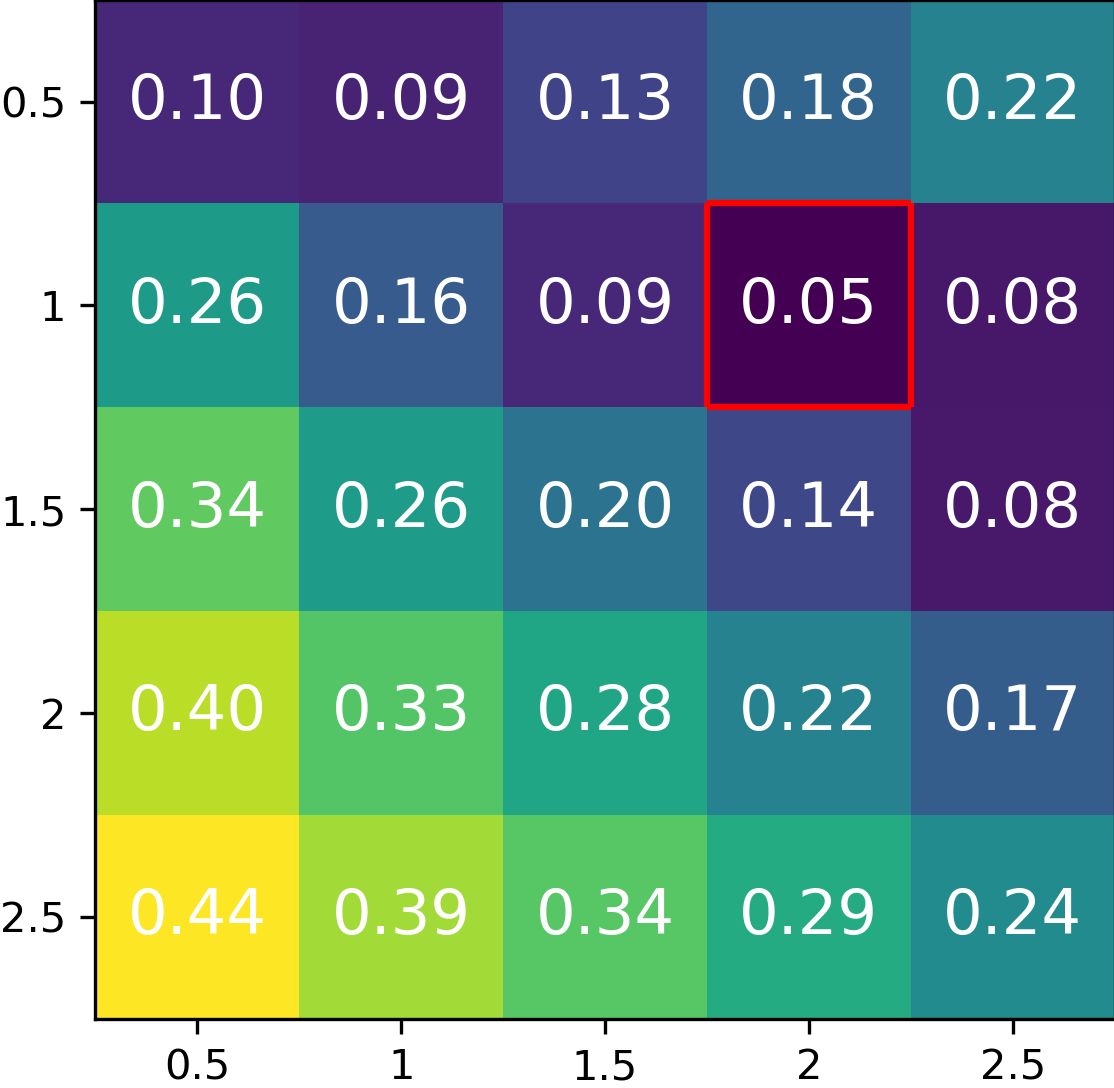}
        \label{fig:2_1}
    }
    \subfigure[s=0.59]{
        \includegraphics[scale=0.35]{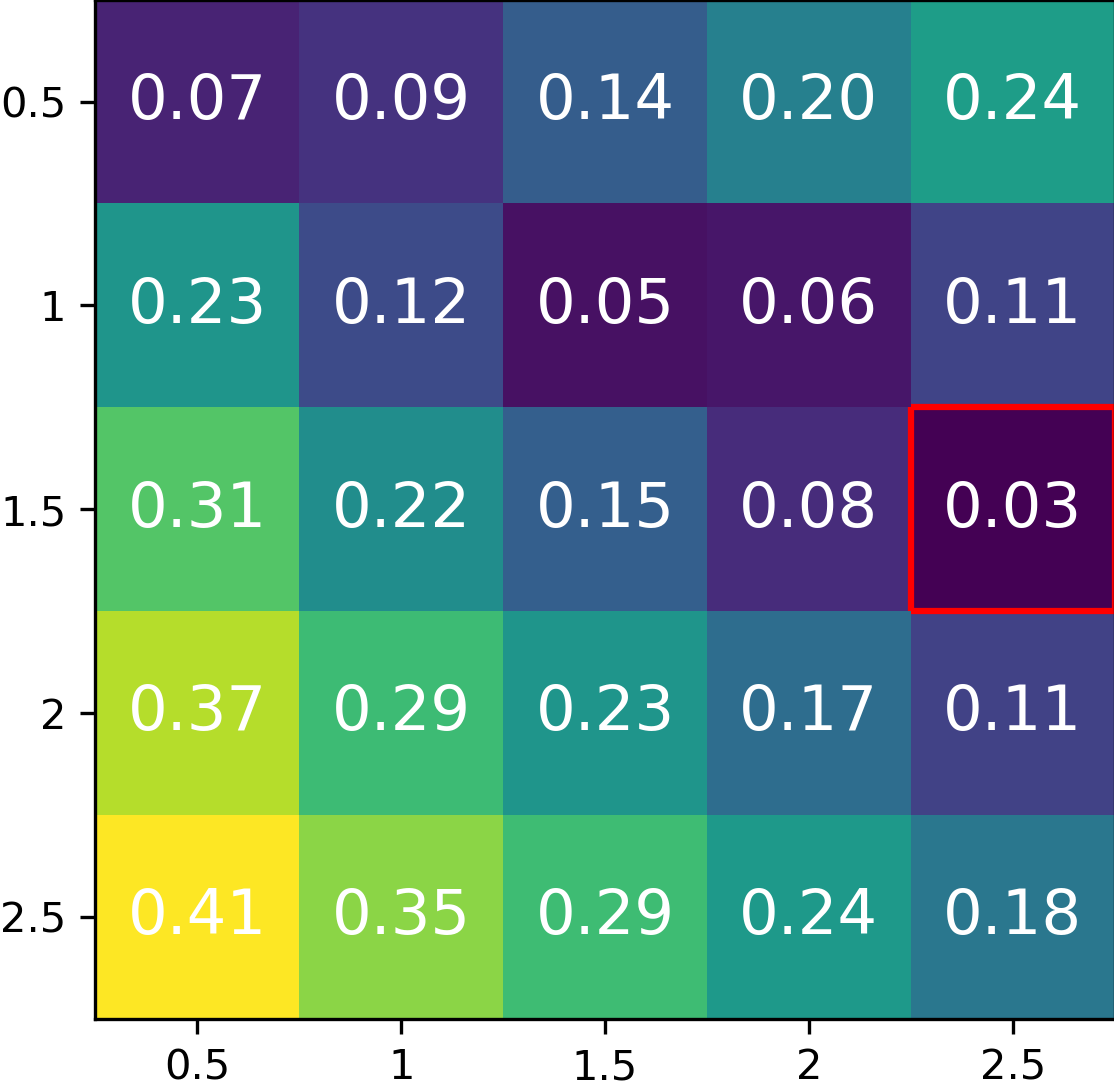}
        \label{fig:2_2}
    }
    \subfigure[s=0.71]{
        \includegraphics[scale=0.35]{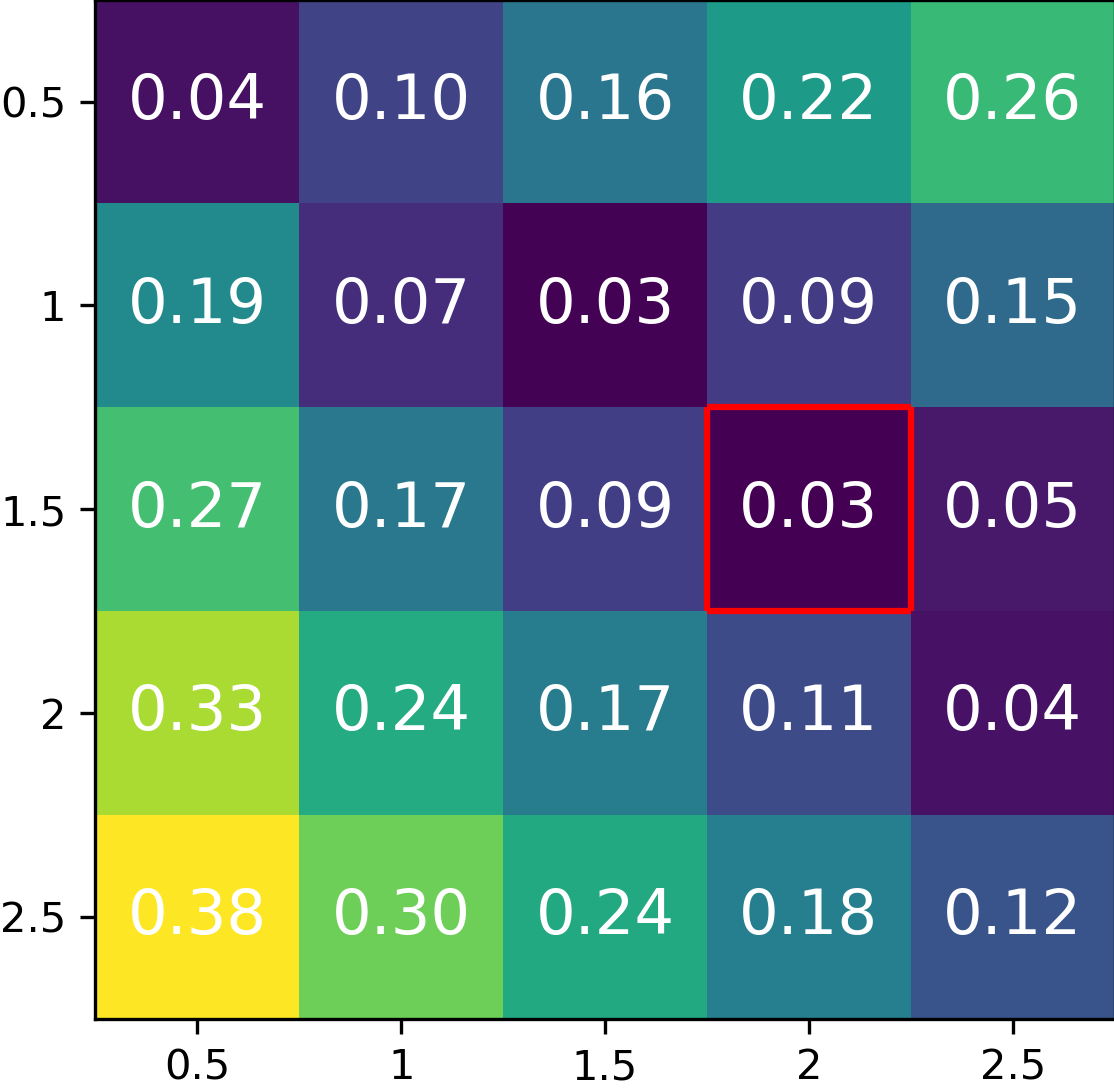}
    }
    \subfigure[s=0.84]{
        \includegraphics[scale=0.35]{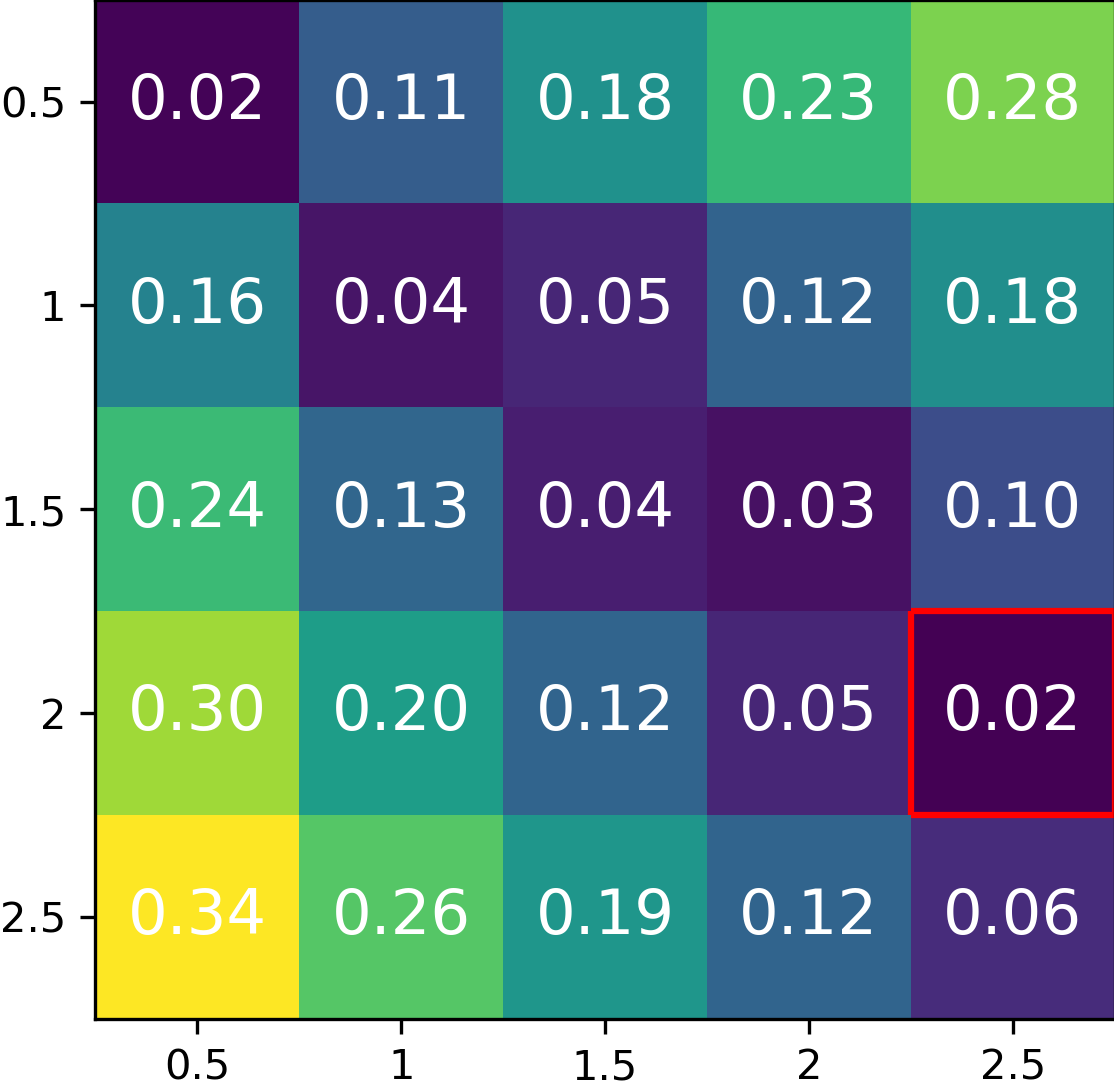}
    }\\
    \subfigure[s=1.19]{
        \includegraphics[scale=0.35]{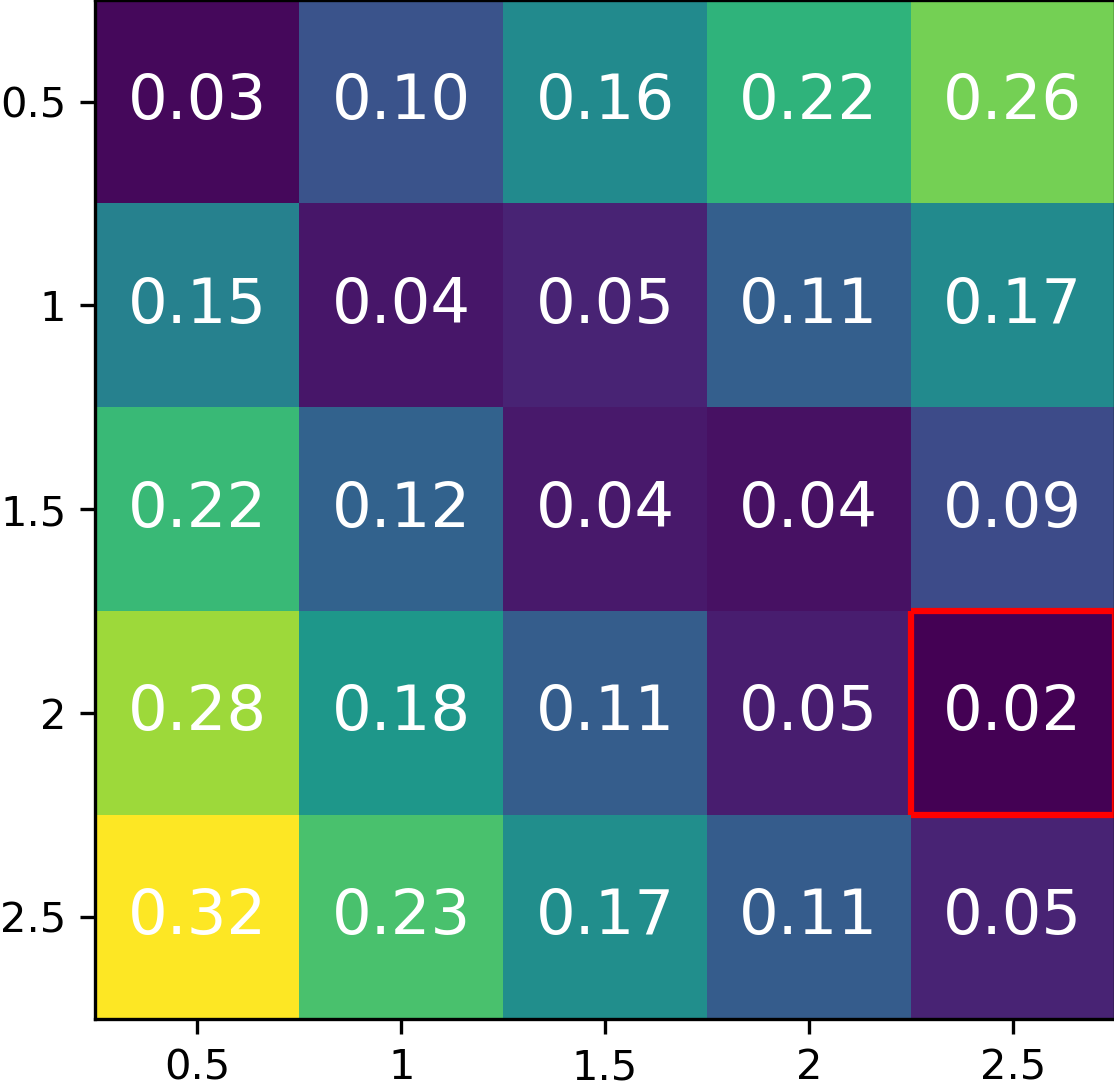}
    }
    \subfigure[s=1.41]{
        \includegraphics[scale=0.35]{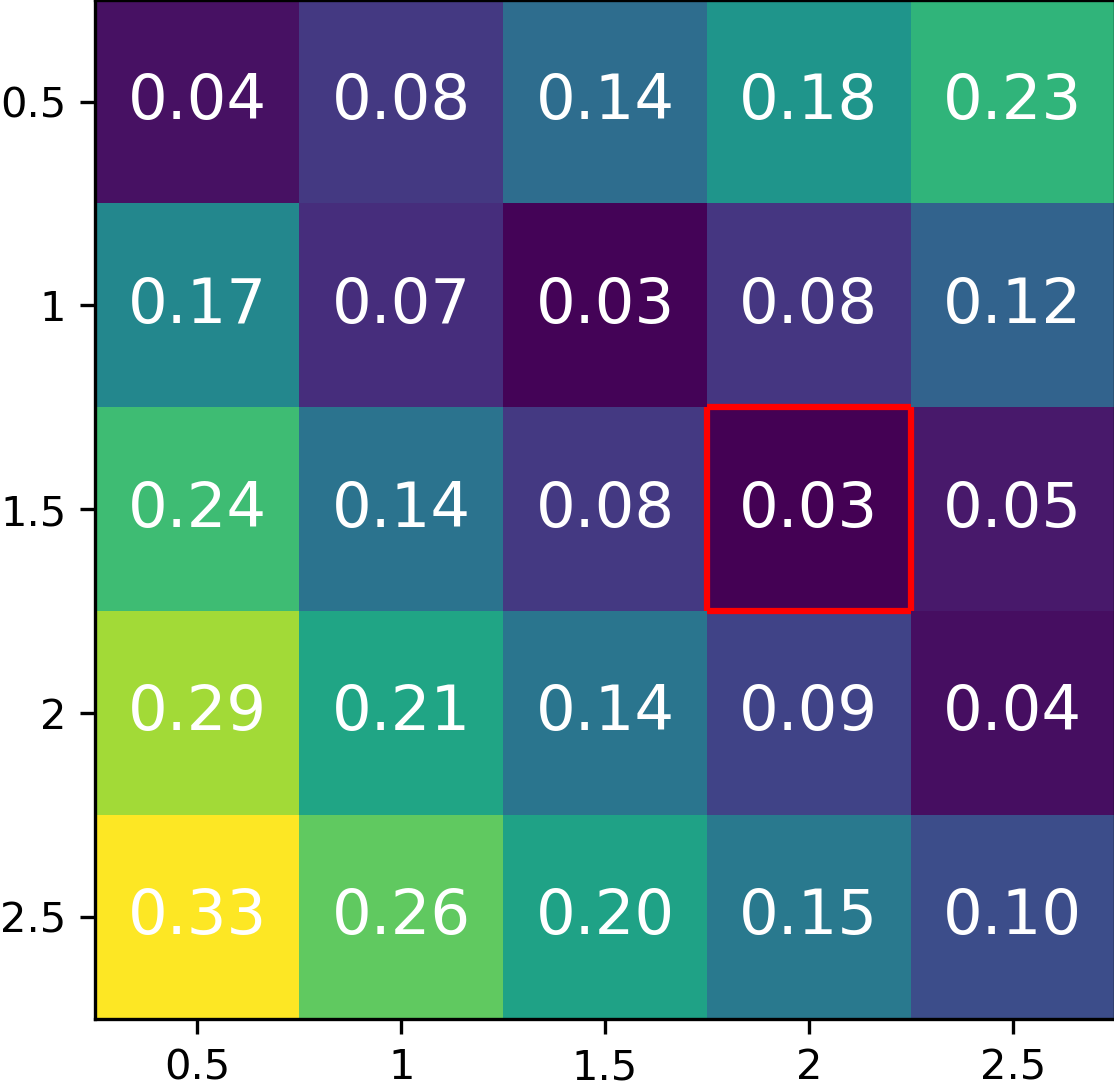}
    }
    \subfigure[s=1.68]{
        \includegraphics[scale=0.35]{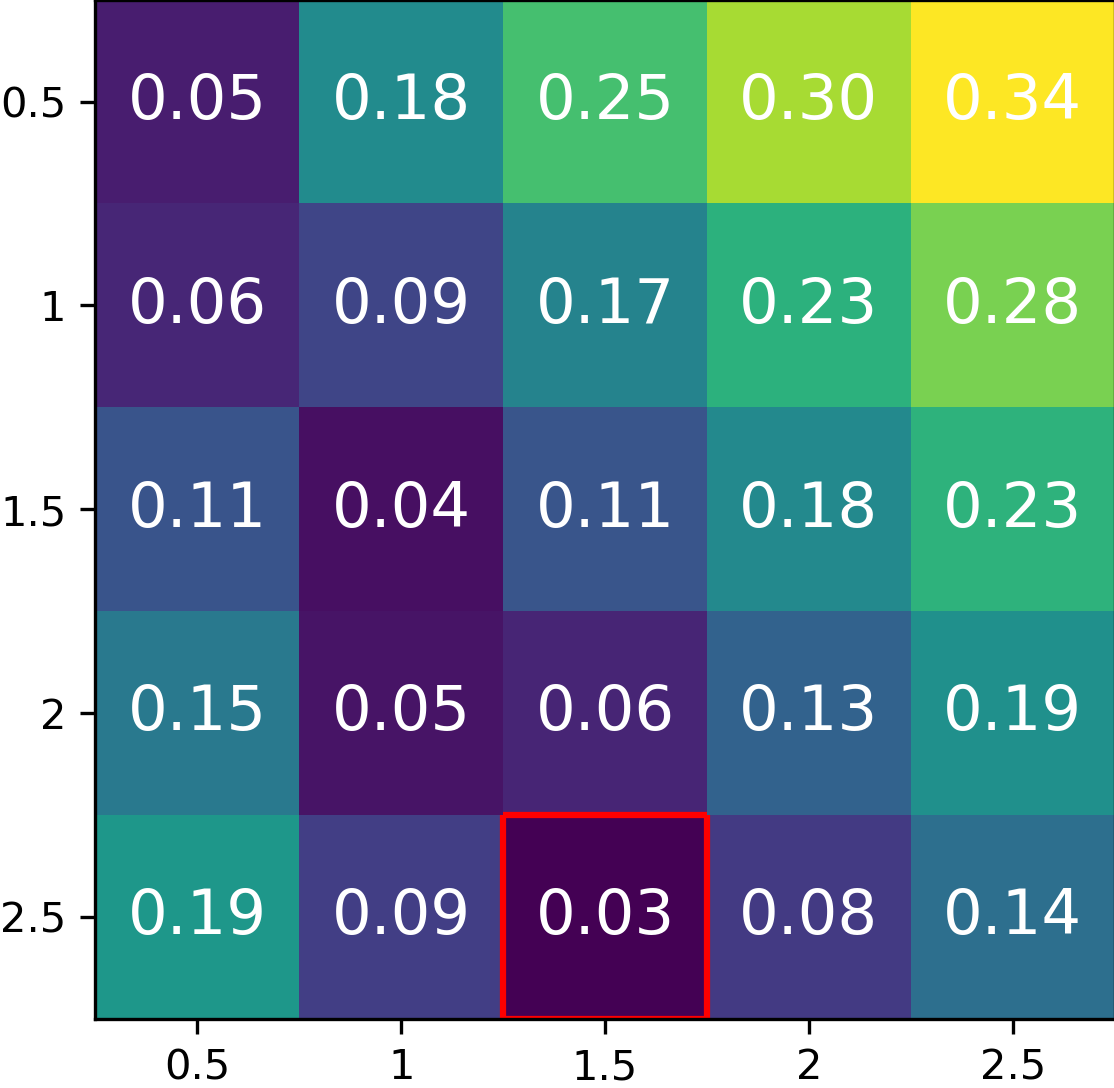}
    }
    \subfigure[s=2.0]{
        \includegraphics[scale=0.35]{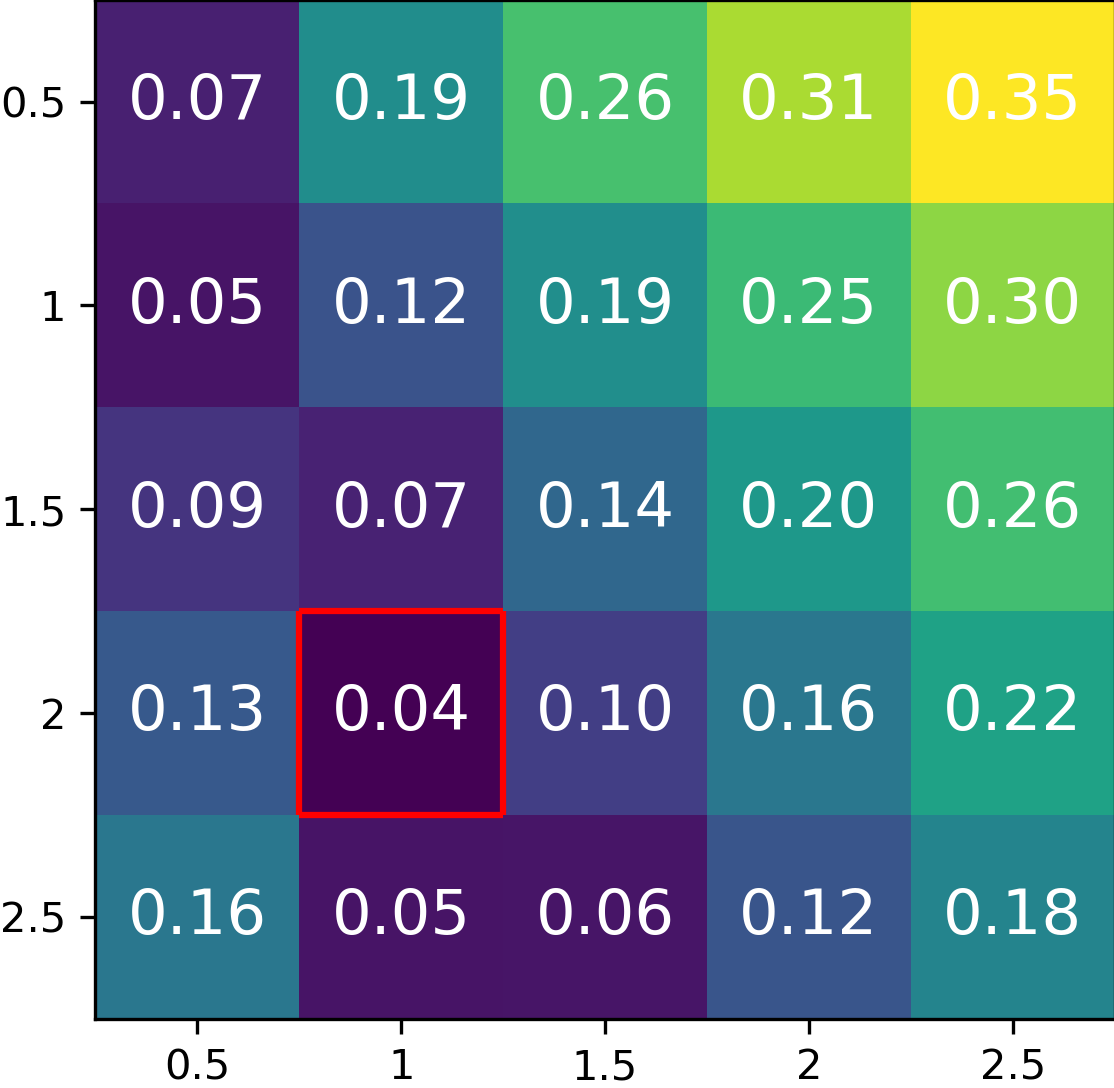}
        \label{fig:3_4}
    }\\
    \subfigure[s=2.38]{
        \includegraphics[scale=0.35]{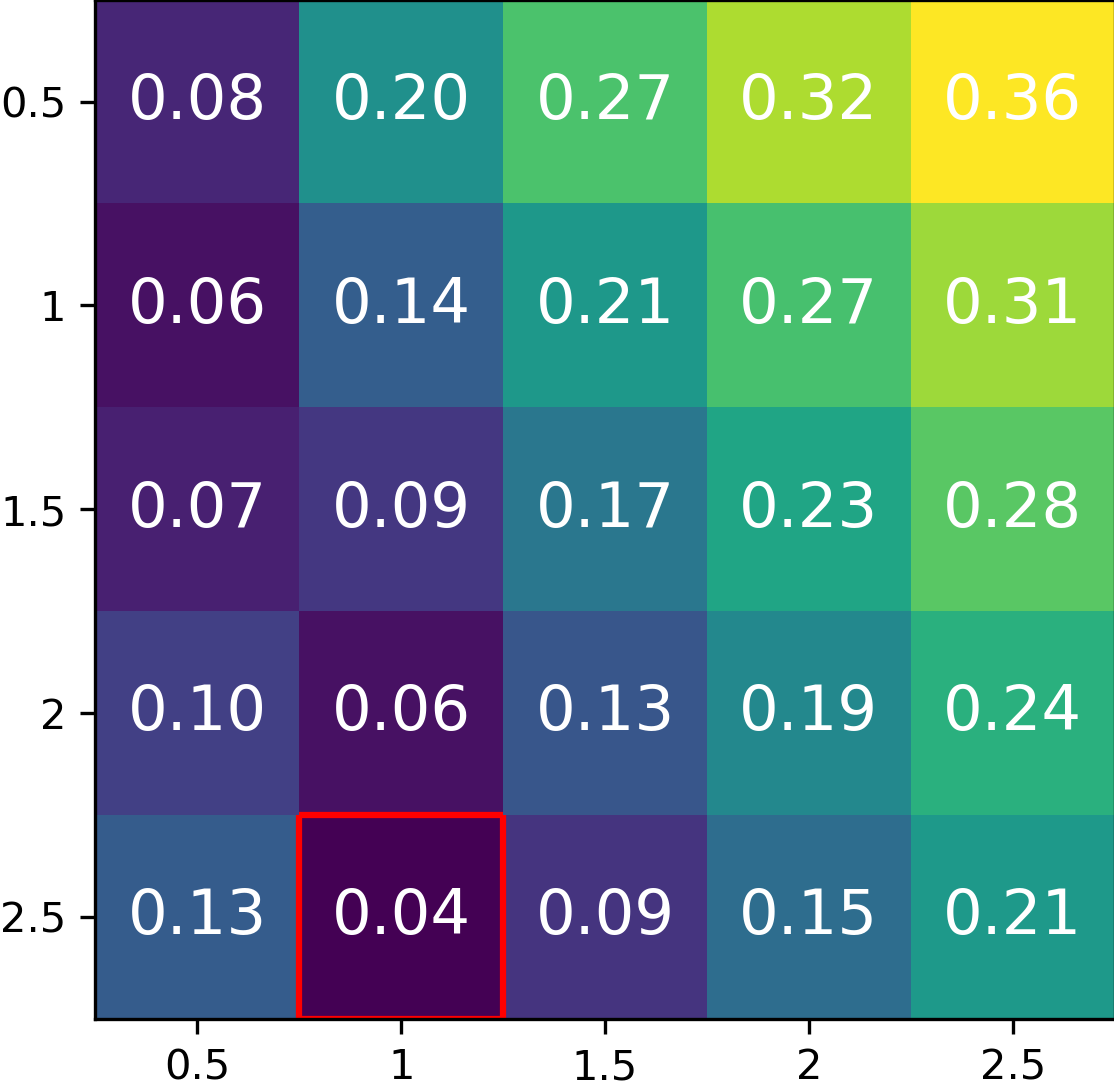}
    }
    \subfigure[s=2.83]{
        \includegraphics[scale=0.35]{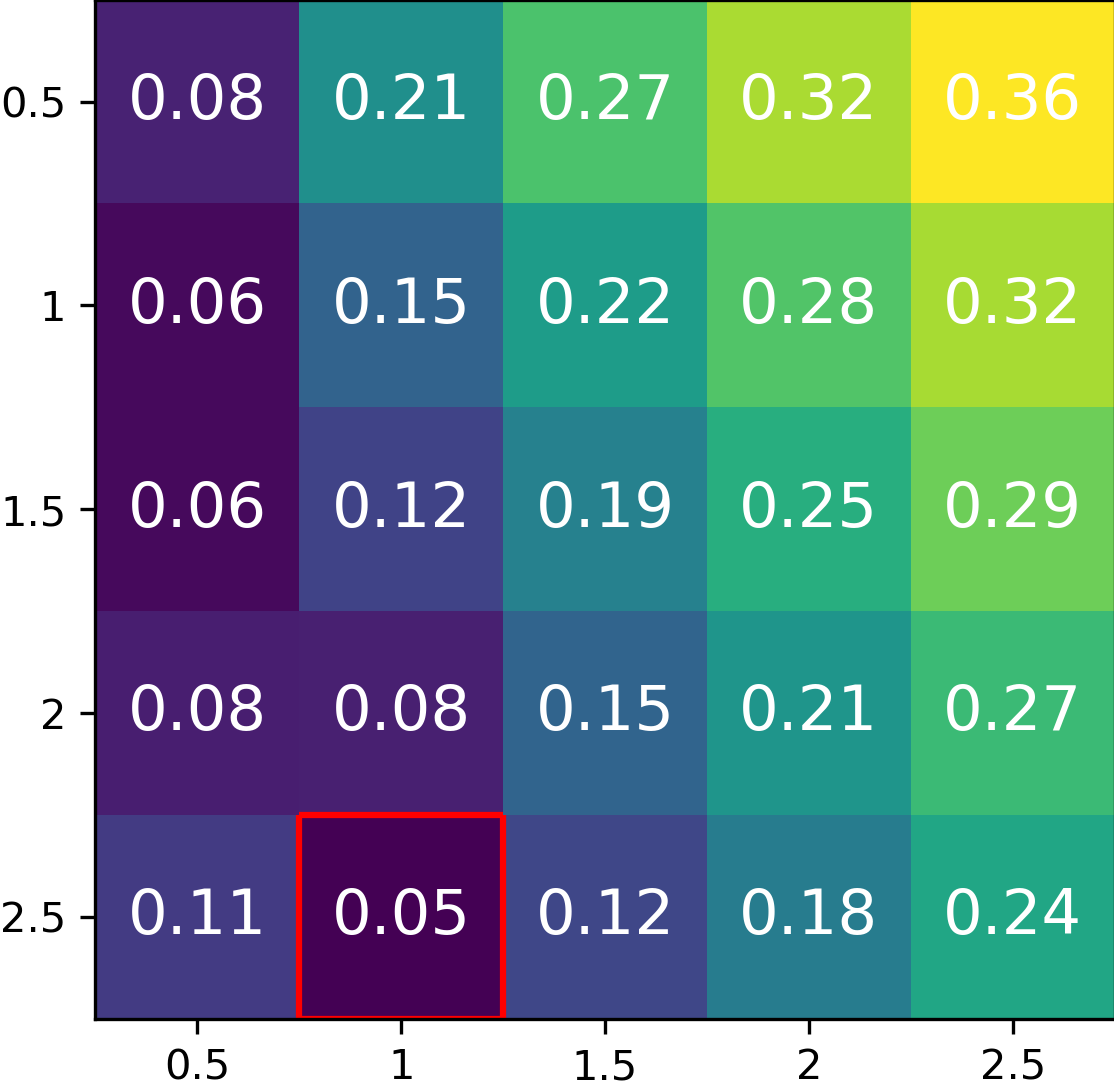}
    }
    \subfigure[s=3.36]{
        \includegraphics[scale=0.35]{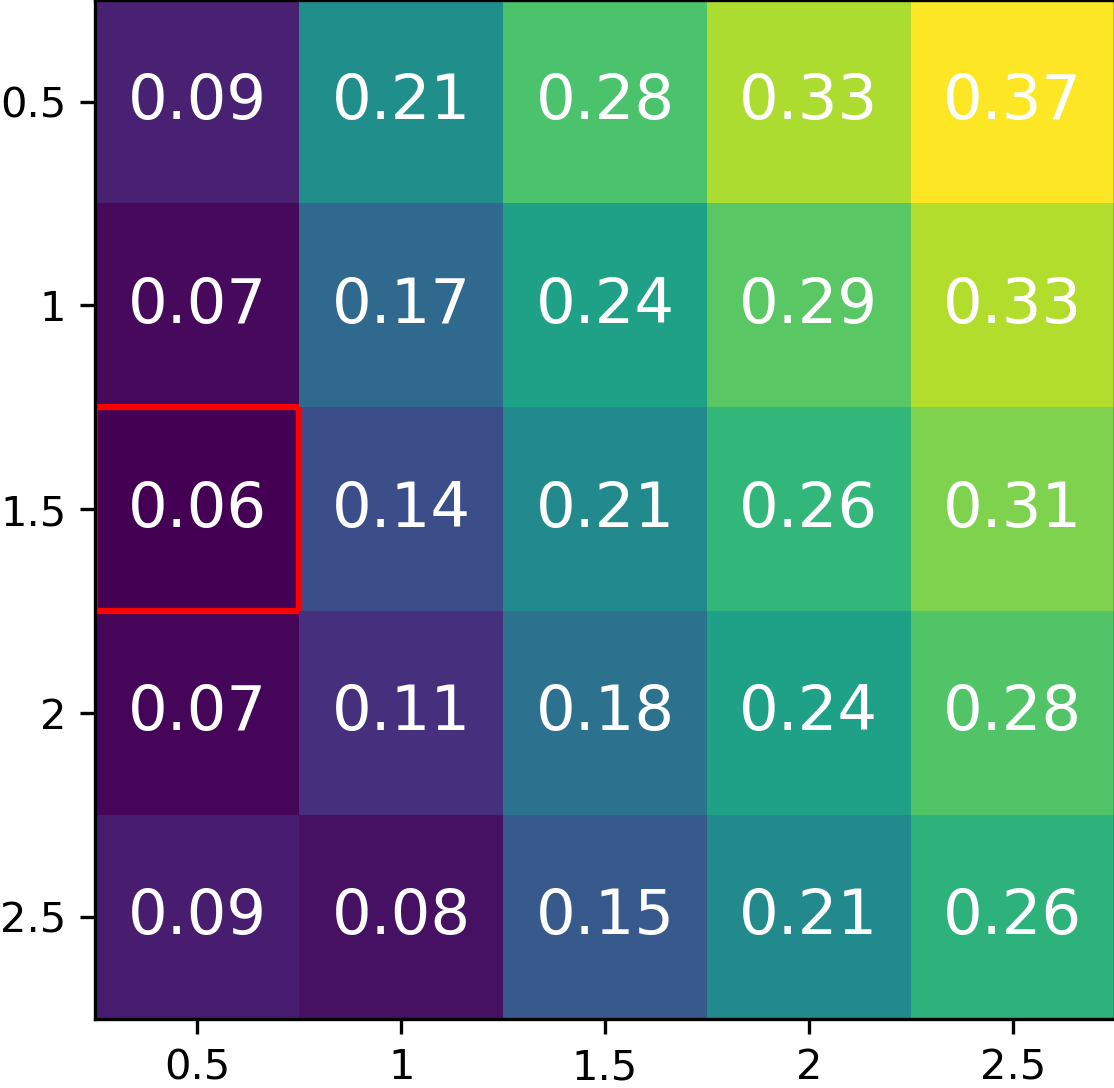}
    }
    \subfigure[s=4.0]{
        \includegraphics[scale=0.35]{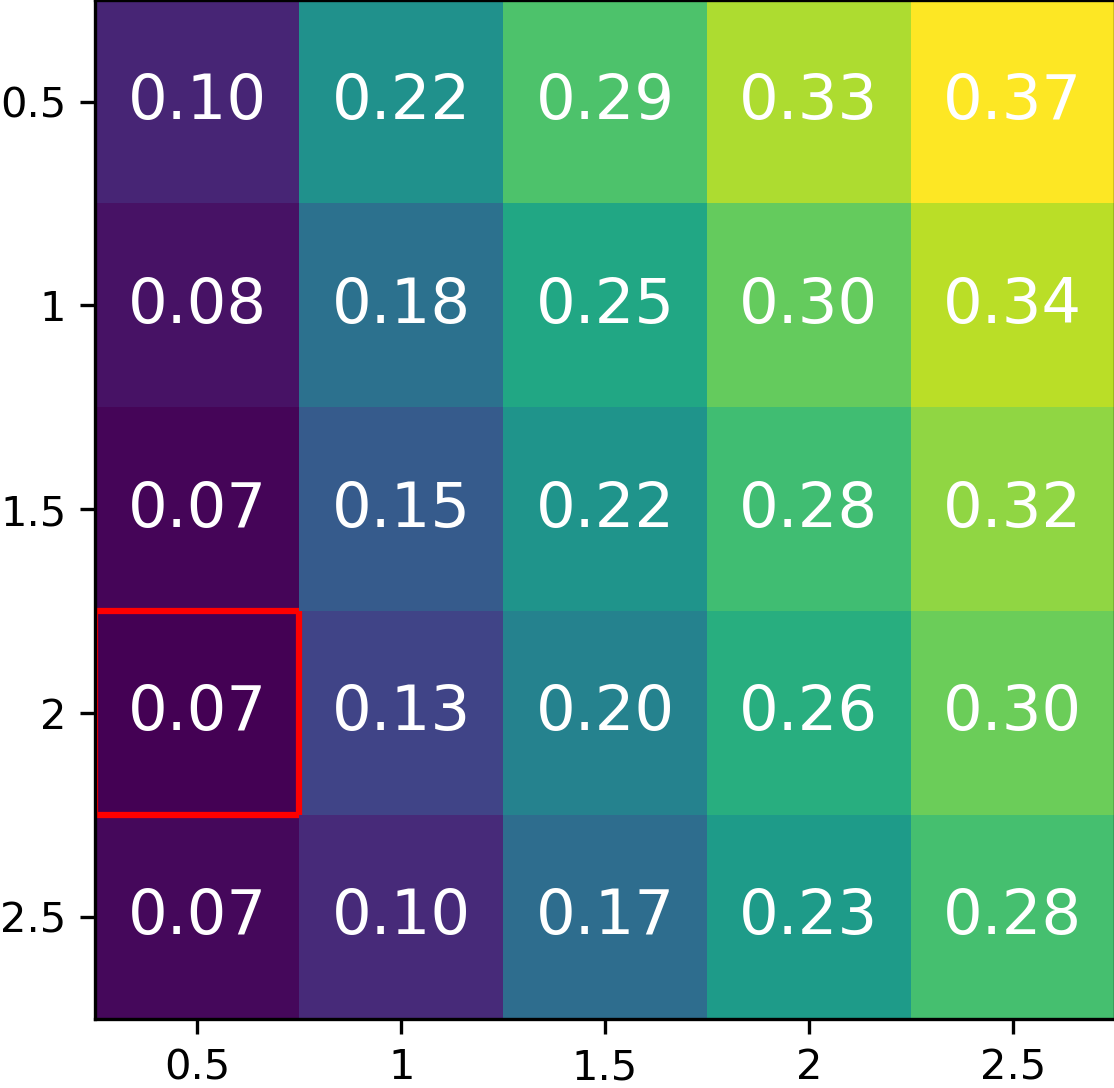}
        \label{fig:4_4}
    }
        
    \caption{Scale equivariance error of feature maps. Each plot shows the equivariance errors between feature maps of the original image and the re-scaled image. In the title of each plot, $s$ denotes the scale factor. The x-axis and the y-axis of each plot denote the $\sigma$ of the filter that is used to convolve with the original image and the re-scaled image, respectively. In each plot, the number on the grid denotes the equivariance error between feature maps $S_s(F_k\star f)$ and $F_{k^\prime}\star S_s(f)$. The lowest equivariance error in each 5$\times$5 error matrix is highlighted by a red box.}
    \label{fig:err}
\end{figure*}

To demonstrate the effectiveness of lowering equivariance error by convolving images with multi-scale filters, we convolve images at different scales with $\gamma$ filters, paring feature maps and then calculate the equivariance error as:
\begin{equation}
    \Delta_{s,k,k\prime} = \frac{1}{N}\sum_{i=1}^N \frac{\|S_s(F_k\star f_i)-F_{k^\prime}\star S_s(f_i)\|_2^2}{\|S_s(F_k\star f_i)\|_2^2},\;s\in\mathbb{R^+},\;k,k^\prime\in\{1,\cdots,\gamma\}
\end{equation}
where $f_i$ is an image, $F_k$ and $F_{k^\prime}$ are filters with scale parameters $\sigma_k$ and $\sigma_{k^\prime}$, $S_s$ is a scaling operation with factor $s$. Thus, given $\gamma$ filters and two images at different scales, we arrive at a $\gamma\times\gamma$ equivariance error matrix. Where each element represents the equivariance error between feature maps, which are obtained by convolving images of different scales with different filters. As shown in Figure~\ref{fig:err}, for the feature map pair that produces the maximal matching, the ratio of scales between images is equal (or close) to the ratio of $\sigma_k$s between filters. For example, in Figure~\ref{fig:4_4}, the ratio between $\sigma$s and the ratio between image scales is the same ($\frac{2}{0.5}=\frac{4}{1}$). The same phenomenon can be observed from images re-scaled by factors of 0.5 and 2 (Figure~\ref{fig:2_1} and~\ref{fig:3_4}). For images whose scales are not divisible, the matching degree between feature maps obtained by convolving the filter with the $\sigma$ ratio closest to the image ratio is the highest. For example, in Figure~\ref{fig:2_2}, the ratio between images ($\frac{1}{0.59}\approx1.69$) is close to the ratio between $\sigma$s ($\frac{2.5}{1.5}\approx1.67$). Thus, we experimentally validated that the scale equivarance err can be reduced by convolving images at different scales with appropriate filters whose scale is corresponded to the scale of images.
\section{Visualisation of Model Prediction}\label{app:visualisation}
\begin{figure*}[!b]
    \centering
    \includegraphics[scale=0.83]{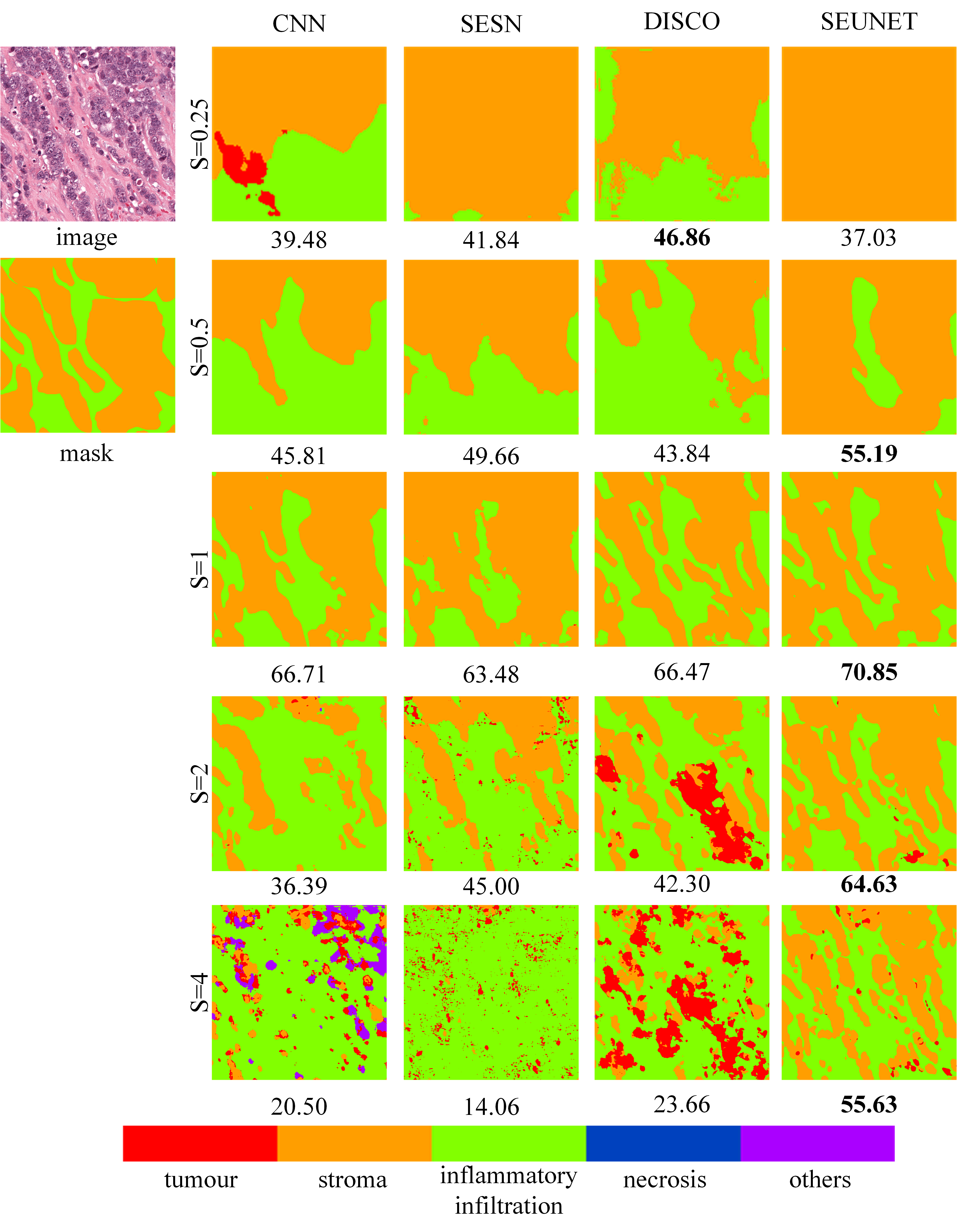}
    \caption{Visual comparison on the BCSS dataset. The mIoU score of each prediction is reported below the segmentation map. The highest score is highlighted in bold. Each column shows segmentation maps of a model on an image re-scaled by different scaling factors ($S=1$ denotes the original scale).}
    \label{fig:bcss_examples}
\end{figure*}

\begin{figure*}
    \centering
    \includegraphics[scale=0.83]{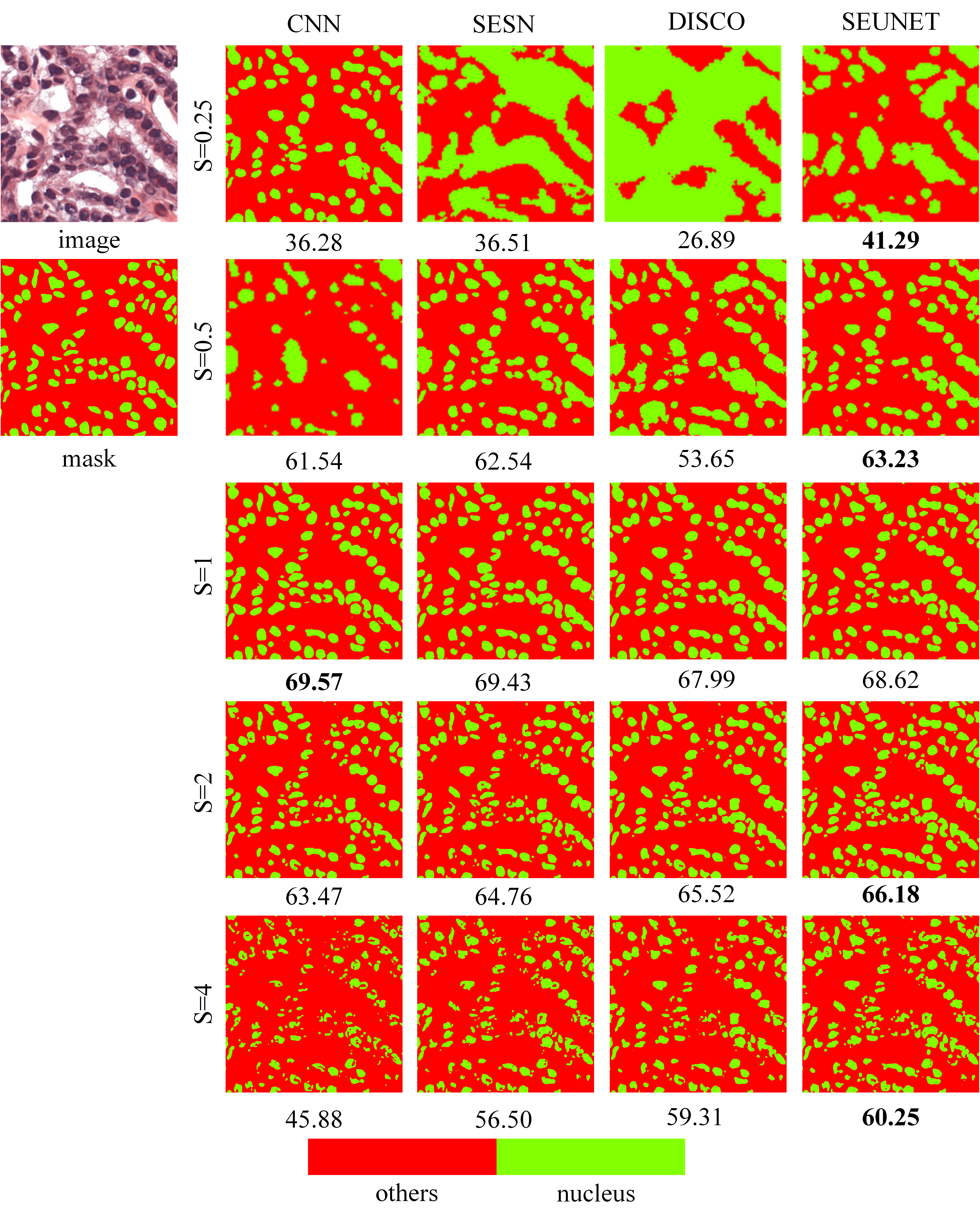}
    \caption{Visual comparison on the MoNuSeg dataset. The IoU score of each prediction is reported below the segmentation map. The highest score is highlighted in bold. Each column shows segmentation maps of a model on an image re-scaled by different scaling factors ($S=1$ denotes the original scale).}
    \label{fig:monuseg_examples}
\end{figure*}

To better understand the SEUNet, we visualise segmentation maps generated by the SEUNet and other compared models on input images at different scales. As shown in Figure~\ref{fig:bcss_examples} and~\ref{fig:monuseg_examples}, the SEUNet can retain a relative decent prediction when compared with other methods. 
\end{document}